%% file: Revisiting_Part1_Main_arXiv.tex
\documentclass[review]{elsarticle}

\usepackage{amssymb}
\usepackage[cmex10]{amsmath}
\usepackage[caption=false,font=normalsize,labelfont=sf,textfont=sf]{subfig}
\usepackage{stmaryrd}
\usepackage{multirow}
\usepackage{tabularx}
\usepackage{booktabs}
\usepackage{algorithm}
\usepackage{algorithmic}
\usepackage{dsfont}
\usepackage[latin1]{inputenc}
\usepackage{array}
\usepackage{longtable}
\usepackage{tabu}
\usepackage{lscape}
\usepackage[pass]{geometry}
\usepackage{mathtools}
\usepackage[title]{appendix}
\usepackage{etoolbox}
\usepackage{adjustbox}
\usepackage{graphicx}
\graphicspath{{}}
\DeclareGraphicsExtensions{.pdf,.jpeg,.png,.eps}

\makeatletter
\def\ps@pprintTitle{%
  \let\@oddhead\@empty
  \let\@evenhead\@empty
  \def\@oddfoot{\reset@font\hfil\thepage\hfil}
  \let\@evenfoot\@oddfoot
}
\makeatother

\newcommand{\E}{\mathrm{E}}
\newcommand{\Prob}{P}

\DeclareMathSymbol{\Delta}{\mathalpha}{operators}{1}

\newcommand{\titlePartI}{Untangling~AdaBoost-based Cost-Sensitive~Classification\protect\\Part I: Theoretical Perspective}
\newcommand{\titlePartII}{Untangling~AdaBoost-based Cost-Sensitive~Classification\protect\\Part II: Empirical Analysis}

\journal{Pattern Recognition}

\begin{document}

\include{Revisiting_Part1}

\section{Acknowledgements}
\label{sec:acknowledgements}{This work has been supported by the Galician Government through the Research Contract GRC2014/024 (Modalidade: Grupos de Referencia Competitiva 2014), and also by the Spanish Government and the European Regional Development Fund (ERDF) under project TACTICA.}

\clearpage
\include{Revisiting_Appendices1}

\bibliography{Revisiting}

\end{document}

%% file: Revisiting_Part1.tex

\begin{frontmatter}

\title{\titlePartI}

\author{Iago Landesa-V\'azquez, Jos\'e Luis Alba-Castro}
\address{Signal Theory and Communications Department, University of Vigo, Maxwell Street, 36310, Vigo, Spain}
\ead{iagolv@gts.uvigo.es, jalba@gts.uvigo.es}

\begin{abstract}
Boosting algorithms have been widely used to tackle a plethora of problems. In the last few years, a lot of approaches have been proposed to provide standard AdaBoost with cost-sensitive capabilities, each with a different focus. However, for the researcher, these algorithms shape a tangled set with diffuse differences and properties, lacking a unifying analysis to jointly compare, classify, evaluate and discuss those approaches on a common basis. In this series of two papers we aim to revisit the various proposals, both from theoretical (Part I) and practical (Part II) perspectives, in order to analyze their specific properties and behavior, with the final goal of identifying the algorithm providing the best and soundest results. 
\end{abstract}

\begin{keyword}
AdaBoost \sep Classification \sep Cost \sep Asymmetry \sep Boosting
\end{keyword}

\end{frontmatter}

\mathtoolsset{showonlyrefs}

\section{Introduction}
\label{sec:Intro1}

The classical approach to solve a classification problem is based on the use of a single expert that must be able to build a solution classifier. However, in the last few decades, a new classification paradigm, based on the combination of several experts in a distributed decision process, has arisen and attracted the attention of the Machine Learning community. The success of this paradigm relies on several theoretical, practical and even biological reasons (such as generalization properties, complexity, data handling, data source fusion, etc.) making these \emph{Ensemble Classifiers} \citep{Polikar06} preferable to classical ones in many scenarios. 

One of the milestones on the history of ensemble methods was the work published by Robert E. Schapire in 1990 \citep{Schapire90}, in which the author proves the equivalence between \emph{weak} learners, algorithms able to generate classifiers performing only slightly better than random guessing, and \emph{strong} learners, those generating classifiers which are correct in all but an arbitrarily small fraction of the instances. This new model of learnability, in which weak learners can be \emph{boosted} to achieve strong performance when they are properly combined, paved the way to one of the most prominent families of algorithms within the ensemble classifiers paradigm: \emph{boosting}.

In 1997, Yoav Freund and Robert E. Schapire \citep{FreundSchapire97} proposed a more general boosting algorithm called AdaBoost (from Adaptive Boosting). Unlike previous approaches, AdaBoost does not require any prior knowledge on weak hypothesis space, and it iteratively adjusts to weak hypothesis that become part of the ensemble. Apart from theoretical guarantees and practical advantages over its predecessors, early experiments on AdaBoost also showed a surprising resistance to overfitting. As a consequence of all these qualities, AdaBoost has received an attention ``rarely matched in computational intelligence'' \citep{Polikar06} being an active research topic in the fields of machine learning, pattern recognition and computer vision \citep{Schapire98, SchapireSinger99,Opitz99,Friedman00,MeaseWyner08a,ViolaJones04,MasnadiVasconcelos11, LandesaAlba12} till present. 

Throughout this time, several studies have been conducted to analyze AdaBoost from different points of view, relating the algorithm with different theories: margin theory \citep{Schapire98}, entropy \citep{Kivinen99}, game theory \citep{FreundSchapire96}, statistics \citep{Friedman00}, etc. In the same way, numerous AdaBoost and boosting variants have been proposed for the two-class and multiclass problems: Real AdaBoost \citep{SchapireSinger99, Friedman00}, LogitBoost \citep{Friedman00}, Gentle AdaBoost \citep{Friedman00}, AsymBoost \citep{ViolaJones02}, AdaCost \citep{Fan99}, AdaBoost.M1 \citep{FreundSchapire96b}, AdaBoost.M2 \citep{FreundSchapire96b}, AdaBoost.MH \citep{SchapireSinger99}, AdaBoost.MO \citep{SchapireSinger99}, AdaBoost.MR \citep{SchapireSinger99}, JointBoosting \citep{Torralba04}, AdaBoost.ECC \citep{GuruswamiSahai99} etc.

Among the different kinds of classification problems, one common subset is that of tasks with clearly different costs depending on each possible decision, or scenarios with very unbalanced class priors in which one class is extremely more frequent or easier to sample than the other one. In such \emph{cost-sensitive} or \emph{asymmetric} conditions (disaster prediction, fraud detection, medical diagnosis, object detection, etc.) classifiers must be able to focus their attention in the rare/most valuable class. Many works in the literature have been devoted to cost-sensitive learning \citep{Elkan01, Provost97, Weiss03}, including a significant set of proposals on how to provide AdaBoost with asymmetric properties (e.g. \citep{Fan99, Ting00, ViolaJones04, ViolaJones02, Sun07, MasnadiVasconcelos07, MasnadiVasconcelos11, LandesaAlba12, LandesaAlba13}). The link between AdaBoost and Cost-Sensitive learning has special interest since AdaBoost is the learning algorithm inside the widespread Viola-Jones object detector framework \citep{ViolaJones04}, a seminal work in computer vision dealing with a markedly asymmetric problem and a enormous number of weak classifiers (the order of hundred of thousands).

The different AdaBoost asymmetric variants proposed in the literature are very heterogeneous, and their related works are focused on emphasizing the possible advantages of each respective method, rather than building a common framework to jointly classify, analyze and discuss the different approaches. The final result is that, for the researcher, these algorithms shape a confusing set with no clear theoretical properties to rule their application in practical problems. 

In this series of two papers we try to classify, analyze, compare and discuss the different proposals on Cost-Sensitive AdaBoost algorithms, in order to gain a unifying perspective. Our final goal is finding a definitive scheme to directly translate any cost-sensitive learning problem to the AdaBoost framework and shedding light on which algorithm can ensure the best performance. 

The current article is focused on the theoretical part of our work and it is organized as follows: next section focuses on standard AdaBoost and its related theoretical framework, Section \ref{sec:CSvar} is devoted to cluster and explain, in an homogeneous notational framework, the different cost-sensitive AdaBoost variants proposed in the literature, and in Section \ref{sec:Discuss} we analyze in depth those algorithms with a fully theoretical derivation scheme. Finally, we present the preliminary conclusions (Section \ref{sec:Conclusions1}) that will be culminated in the accompanying paper ``\titlePartII'' covering the experimental part of our work.

\section{AdaBoost}
\label{sec:AdaBoost}


Let us define $\mathbf{X}$ as the random process from which our observations $\mathbf{x}=\left(x_{1},\ldots,x_{N}\right)^{T}$ are sampled, and $Y$ the random variable governing the related labels $y \in \{-1,1\}$. In this scenario, a \emph{detector} $H(\mathbf{x})$ (we will also refer to it as \emph{classifier} or  \emph{hypothesis}) is a function trying to guess the label $y$ of a given sample $\mathbf{x}$, and it can be defined in terms of a more generic function $f\left(\mathbf{x}\right) \in \mathbb{R}$ which we will call \emph{predictor}.

\begin{equation}
\label{pred_eqn}
H(\mathbf{x})=\mathrm{sign}\left[f\left(\mathbf{x}\right)\right]
\end{equation}

Suppose we have a training set of $n$ examples $\mathbf{x}_i$ with its respective labels $y_i$, a weight distribution $D(i)$ over them and a \emph{weak learner} able to select, according to labels and weights, the best detector $h(\mathbf{x})$ from a predefined collection of weak classifiers. In this scenario, the role of AdaBoost is to compute a goodness measure $\alpha$ depending on the performance obtained by the selected weak classifier, and to update, accordingly, the weight distribution to emphasize misclassified training examples. Then, with a different weight distribution, the weak learner can make a new hypothesis selection and the process restarts. By iteratively repeating this scheme (\refeq{alphat_eqn}, \refeq{rt_eqn}, \refeq{weight_rule_eqn}, \refeq{zt_eqn}) with $t$ indexing the number of learning rounds, AdaBoost obtains an ensemble of weak classifiers with respective goodness parameters $\alpha_{t}$. 

\begin{equation}
\label{alphat_eqn}
\alpha_{t}=
\frac{1}{2} \log \left(\frac{1+r_{t}}{1-r_{t}}\right)
\end{equation}

\begin{equation}
\label{rt_eqn}
r_{t}=
\sum_{i=1}^{n}D_{t}(i) y_{i} h_{t} (\mathbf{x}_{i})
\end{equation}

\begin{equation}
\label{weight_rule_eqn}
D_{t+1}(i)=
\frac{D_{t}(i)\exp\left(-\alpha_{t} y_{i} h_{t}(\mathbf{x}_{i})\right)}{Z_{t}}
\end{equation}

\begin{equation}
\label{zt_eqn}
Z_{t}=
\sum_{i=1}^{n} D_{t}(i) \exp\left(-\alpha_{t} y_{i} h_{t}(\mathbf{x}_{i})\right)
\end{equation}

Weak hypothesis searching in AdaBoost is guided to maximize goodness $\alpha_{t}$ of each selected classifier, which is equivalent to maximize, at each iteration, weighted correlation $r_{t}$ (\refeq{rt_eqn}) between labels $y_{i}$ and predictions $h_{t}$. This iterative searching process can continue until a predefined number $T$ of training rounds have been completed or some performance goal is reached. The final AdaBoost \emph{strong detector} $H(\mathbf{x})$ is defined (\refeq{stradb_eqn}) in terms of a boosted \emph{predictor} $f(\mathbf{x})$ built as an ensemble of the selected weak classifiers weighted by their respective goodness parameters $\alpha_{t}$.

\begin{equation}
\label{stradb_eqn}
H(\mathbf{x})=\mathrm{sign}\left(f(\mathbf{x})\right)=\mathrm{sign}\left(\sum_{t=1}^{T}\alpha_{t}h_{t}(\mathbf{x})\right)
\end{equation}

\subsection{Error Bound Minimization}
\label{subsec:GeneralizedVersion}

Robert E. Schapire and Yoram Singer proposed \citep{SchapireSinger99}, from the original derivation of AdaBoost, a generalised and simplified analysis that models the algorithm as an additive (round-by-round) minimization process of an exponential bound on the strong classifier training error ($E_{T}$). This bounding process is explained in equation\footnote{Notation: $\llbracket a \rrbracket$ is $1$ when $a$ is true and $0$ otherwise.} (\refeq{bound_ineq_eqn}) from which all AdaBoost equations we have presented, weight update rule included, can be derived \citep{LandesaAlba12}.

\begin{equation}
\label{bound_ineq_eqn} 
{\begin{array}{c} 
\underbrace{\strut H(\mathbf{x}_{i})\neq y_{i} \:\Rightarrow\: y_{i} f(\mathbf{x}_{i}) \leq 0 \:\Rightarrow\: \exp\left(-y_{i} f(\mathbf{x}_{i})\right) \geq 1}\\
\Downarrow\\
E_{T}= \sum_{i=1}^{n} D_{1}(i) \llbracket H(\mathbf{x}_{i}) \neq y_{i}\rrbracket \leq \sum_{i=1}^{n} D_{1}(i) \exp \left( -y_{i} f(\mathbf{x}_{i}) \right)
\end{array}
}
\end{equation}

After (\refeq{bound_ineq_eqn}), the final bound of the training error obtained by AdaBoost can be expressed as (\refeq{et_bound_eqn}), and the additive minimization of the exponential bound $\tilde{E}_{T}$ can be seen as finding, in each round, the weak hypothesis $h_{t}$ that maximizes $r(t)$, the weighted correlation between labels $(y_{i})$ and predictions $(h_{t})$. 

\begin{equation}
\label{et_bound_eqn}
E_{T}\leq
\prod_{t=1}^{T}Z_{t}\leq
\prod_{t=1}^{T}\sqrt{1-{r_{t}}^2}=\tilde{E}_{T}
\end{equation}

When weak hypothesis are binary, $h_{i}\in\{-1,+1\}$, the last inequality on (\refeq{et_bound_eqn}) becomes an equality, and parameter $\alpha_{t}$ can be directly rewritten (\refeq{alphat2_eqn}) in terms of the weighted error $\epsilon_{t}$ of the current weak classifier. As can be seen, the minimization process turns out to be equivalent to simply selecting the weak classifier with less weighted error.

\begin{equation}
\label{round_err_eqn}
\epsilon_{t}=
\sum_{i=1}^{n} D_{t}(i) \llbracket h_{t}(\mathbf{x}_{i}) \neq y_{i}\rrbracket=
\sum_{\textrm{err}}D_{t}(i)
\end{equation}

\begin{equation}
\label{alphat2_eqn}
\alpha_{t}=
\frac{1}{2} \log \left( \frac{1-\epsilon_{t}}{\epsilon_{t}}\right)
\end{equation}

In line with other works, for the sake of simplicity and clarity, we will focus our analysis on this \emph{Discrete} version of AdaBoost using binary weak classifiers, which does not prevent our conclusions from being extended to other variations of the algorithm. Also, trying to define an homogeneous notational framework for our work, we have unified the different notations found in the literature to that used by Schapire and Singer \citep{SchapireSinger99}. A summary of AdaBoost can be found on Algorithm \ref{adb_algorithm} (all the algorithms discussed in this paper are detailed, with homogeneous notation, in Appendix \ref{app:algorithms}).

\subsection{Statistical View of Boosting}
\label{subsec:Statistical View}

One of the milestones in boosting research and the foundation of many variations of AdaBoost is the highly-cited contribution by Jerome H. Friedman et al. \citep{Friedman00} in which a statistical reinterpretation of boosting is given. Following the exponential criterion seen in the last subsection, Friedman et al. showed that AdaBoost can be motivated as an iterative algorithm building an additive logistic regression model $f(\mathbf{x})$ that minimizes the expectation of the exponential loss, $J(f(\mathbf{x}))$:

\begin{equation}
\label{loss_eqn}
J(f(\mathbf{x}))=\E\left[\exp(-yf(\mathbf{x}))\right]
\end{equation}

This defined loss is effectively minimized at 
\begin{equation}
\label{regress_eqn}
f(\mathbf{x})=\frac{1}{2}\log\left(\frac{\Prob(y=1|\mathbf{x})}{\Prob(y=-1|\mathbf{x})}\right)
\end{equation}
so a direct connection between boosting and additive logistic regression models is drawn. According to this statistical perspective, AdaBoost predictions can be seen as estimations of the posterior class probabilities, which has served as basis to develop many extensions and variants of the algorithm (among them, the Cost-Sensitive Boosting scheme \citep{MasnadiVasconcelos07, MasnadiVasconcelos11}). 

It is important to mention that, despite the huge and unquestionable value of the statistical view, some enriching controversy, revealed by empirical evidences \citep{MeaseWyner08a, Bennet08, MeaseWyner08b}, has arisen about inconsistencies of this interpretation.

\section{Cost-Sensitive Variants of AdaBoost}
\label{sec:CSvar}


Cost-sensitive classification problems can be fully portrayed by a cost matrix \citep{Elkan01} whose components map the loss of each possible result. For two-class problems there are four kinds of results: true positives, true negatives, false positives and false negatives; so the cost matrix $\mathbf{C}$ can be defined as follows: 

\begin{equation}
\label{cost_matrix}
\begin{tabular} {c c c c c c l l l}
&  & \multicolumn{2}{c}{Actual} & & & \\
&  &  Negative & Positive & & &\\
\multirow{2}{*}{$\mathbf{C}=$} & \multirow{2}{*}{{\Huge(}} & $c_{nn}$ & $c_{np}$ & \multirow{2}{*}{{\Huge)}} & Negative & \multirow{2}{*}{Classified}\\
&  & $c_{pn}$ & $c_{pp}$ & & Positive & &
\end{tabular}
\end{equation}

The optimal decision for a given cost matrix will not change if all its coefficients are added a constant, or if they are multiplied by a constant positive factor. As a result, a cost matrix for two-class classification problems only has two degrees of freedom and can be parametrized by only two coefficients: false negatives normalized cost ($\overline{c}_{np}$) and true positives normalized cost ($\overline{c}_{pp}$):

\begin{equation}
\label{simple_C}
\mathbf{C}=\left(
\begin{array} {c c}
0 & \overline{c}_{np}\\
1 & \overline{c}_{pp}\\
\end{array}
\right)
\end{equation}

In the most common case correct decisions have null related costs ($c_{nn}=c_{pp}=0$), so $\mathbf{C}$ has eventually only one degree of freedom: the ratio between cost of errors on positives ($c_{np}$) and cost of errors on negatives ($c_{pn}$). In the literature and most practical problems, cost requirements are usually specified by these two error parameters, which, for simplicity, we will denote as $C_P$ and $C_N$ respectively. 

\begin{equation}
\label{simple_C2}
\mathbf{C}=\left(
\begin{array} {c c}
0 & C_P/C_N\\
1 & 0\\
\end{array}
\right)
\rightarrow \left(
\begin{array} {c c}
0 & C_P\\
C_N & 0\\
\end{array}
\right)
\end{equation}


The coefficients of a cost matrix may not be constant in general. While constant coefficients model a scenario where all the examples of each class have the same cost (class-level asymmetry), variable coefficients mean that examples belonging to the same class can have different costs (example-level asymmetry). Whatever the scenario, it is also important to notice that, for ``reasonableness'' \citep{Elkan01}, correct predictions in a cost matrix should have lower associated costs than mistaken ones ($c_{nn}<c_{np}$ and $c_{pp}<c_{pn}$).

Bearing in mind that class-level asymmetry is the most common for detection problems, and that example-level asymmetry can be modeled by a class-level asymmetry scheme with a resampled training dataset, for our analysis we have homogenized the different Asymmetric AdaBoost approaches to the class-level scheme. Thus, we will follow a prototypical cost-sensitive detection statement specified by two constant coefficients $C_P$ and $C_N$, that can be alternatively described by the ``normalized cost asymmetry'' of the problem $\gamma \in (0,1)$:

\begin{equation}
\gamma=\frac{C_{P}}{C_{P}+C_{N}}
\end{equation}

Despite the widespread use of these particularizations, in Appendix \ref{app:cost_scen} we will extend our conclusions to example-level asymmetry and also cases in which correct classification costs are nonzero.

It is also important to emphasize that this work is focused on AdaBoost and its cost-sensitive variants, a realm of methods in the literature that are based on a exponential loss minimization criterion, analogous to that giving rise to the original algorithm (as we have seen in Sections \ref{subsec:GeneralizedVersion} and \ref{subsec:Statistical View} from different points of view) . Other boosting algorithms based on other kinds of losses beyond the exponential paradigm, like the binomial log-likelihood \citep{Friedman00} or the p-norm loss \citep{LozanoAbe08}, are outside the scope of the current study.

\subsection{Classification}
\label{subsec:Classification}

In order to give a clear overview of the cost-sensitive variants of AdaBoost proposed in the literature, we suggest an analytical classification scheme to cluster them into three categories according to the way asymmetry is reached: \emph{A posteriori}, \emph{Heuristic} and \emph{Theoretical}.

\subsubsection{A Posteriori}
\label{subsec:APosteriori}

The seminal face/object detector framework by Paul Viola and Michael J. Jones \citep{ViolaJones04} uses a validation set to modify, after training, the threshold of the original (cost-insensitive) AdaBoost strong classifier. The goal is to adjust the balance between false positive and detection rates, building, that way, a cost-sensitive boosted classifier:

\begin{equation}
\label{vjmod_eqn}
\tilde{H}(\mathbf{x})=\mathrm{sign}\left(f(\mathbf{x})-\phi\right)=\mathrm{sign}\left({\sum_{t=1}^{T}\alpha_{t}h_{t}(\mathbf{x})}-\phi\right)
\end{equation}

Besides the great success of the detection framework, the authors themselves acknowledge that neither this a posteriori cost-sensitive tuning ensures that the selected weak classifiers are optimal for the asymmetric goal \citep{ViolaJones02}, nor their modifications preserve the original AdaBoost training and generalization guarantees \citep{ViolaJones04}.

An useful insight on this can be drawn from the analysis by Masnadi-Shirazi and Vasconcelos \citep{MasnadiVasconcelos11}. According to the Bayes Decision Rule, the optimal predictor $f^{*}(\mathbf{x})$ can be expressed in terms of the optimal predictor for a cost-insensitive scenario $f^{*}_{0}(\mathbf{x})$ and a threshold $\phi$ depending on costs. 

\begin{equation}
\label{pred_thresh}
f^{*}\left(\mathbf{x}\right)=\log\left(\frac{\Prob_{Y|\mathbf{X}}\left(1|\mathbf{x}\right)C_{P}}{\Prob_{Y|\mathbf{X}}\left(-1|\mathbf{x}\right)C_{N}}\right)=\log\left(\frac{\Prob_{Y|\mathbf{X}}\left(1|\mathbf{x}\right)}{\Prob_{Y|\mathbf{X}}\left(-1|\mathbf{x}\right)}\right)-\log\left(\frac{C_N}{C_P}\right)=f^*_0(\mathbf{x})-\phi
\end{equation}

As a consequence, for any cost requirements, the optimal cost-sensitive \emph{detector} $H^{*}(\mathbf{x})$ can also be expressed as a threshold on the cost-insensitive optimal \emph{predictor} $f^{*}_{0}(\mathbf{x})$.

\begin{equation}
\label{optimal_detector}
H^*\left(\mathbf{x}\right)=\mathrm{sign}\left[f^*(\mathbf{x})\right]=\mathrm{sign}\left[f^*_0(\mathbf{x})-\phi\right]
\end{equation}

In practical terms, however, learning algorithms do not have access to the exact probability distributions and they must approximate this optimal detector rule. Thus, AdaBoost can be seen as an algorithm obtaining an approximation ($\hat{H}_0(\mathbf{x})$) to the optimal cost-insensitive \emph{detector}, built by means of an estimation ($\hat{f}_0(\mathbf{x})$) of the cost-insensitive \emph{predictor} (\refeq{adaboost_estimation}).

\begin{equation}
\label{adaboost_estimation}
\hat{H}_0(\mathbf{x})=\mathrm{sign}\left[\hat{f}_0(\mathbf{x})\right]=\mathrm{sign}\left({\sum_{t=1}^{T}\alpha_{t}h_{t}(\mathbf{x})}\right) \approx  H^*_0(\mathbf{x}) 
\end{equation}

By definition, the purpose of AdaBoost is to obtain a \emph{detector} as close as possible to the optimal one, and this optimality is ensured if the learned \emph{predictor} satisfies two necessary and suficient conditions:

\begin{equation}
\label{costinsens_cond}
{\begin{array}{c} 
\hat{H}_0(\mathbf{x})=H_0^*(\mathbf{x}) \\
\Updownarrow \\
\begin{cases}
\hat{f}_0\left(\mathbf{x}\right)=f_0^*(\mathbf{x})=0 & \text{if } \Prob_{Y|\mathbf{X}}(1|\mathbf{x})=\Prob_{Y|\mathbf{X}}(-1|\mathbf{x})\\
\mathrm{sign}\left[\hat{f}_0\left(\mathbf{x}\right)\right]=\mathrm{sign}\left[f_0^*(\mathbf{x})\right] & \text{if } \Prob_{Y|\mathbf{X}}(1|\mathbf{x}) \neq \Prob_{Y|\mathbf{X}}(-1|\mathbf{x})
\end{cases}
\end{array}
}
\end{equation}

As can be seen, in order to reach optimal \emph{detection} the predictor learned by AdaBoost should match the optimal predictor in the boundary region, but only its sign elsewhere. Analogously, optimal detection for the cost-sensitive case, would be ensured by two equivalent conditions:

\begin{equation}
\label{costsens_cond}
{\begin{array}{c} 
\hat{H}(\mathbf{x})=H^*(\mathbf{x}) \\
\Updownarrow \\
\begin{cases}
\hat{f}\left(\mathbf{x}\right)=f^*(\mathbf{x})=0 & \text{if } \Prob_{Y|\mathbf{X}}(1|\mathbf{x})C_P=\Prob_{Y|\mathbf{X}}(-1|\mathbf{x})C_N\\
\mathrm{sign}\left[\hat{f}\left(\mathbf{x}\right)\right]=\mathrm{sign}\left[f^*(\mathbf{x})\right] & \text{if } \Prob_{Y|\mathbf{X}}(1|\mathbf{x})C_P \neq \Prob_{Y|\mathbf{X}}(-1|\mathbf{x})C_N
\end{cases}
\end{array}
}
\end{equation}

Thus, optimality conditions required by the \emph{a posteriori} modification of the AdaBoost threshold would be as follows:

\begin{equation}
\label{threshmod_cond}
{\begin{array}{c} 
\hat{H}(\mathbf{x})=H^*(\mathbf{x}) \\
\Updownarrow \\
\begin{cases}
\hat{f}_0\left(\mathbf{x}\right)=f_0^*(\mathbf{x})=\phi & \text{if } \Prob_{Y|\mathbf{X}}(1|\mathbf{x})C_P=\Prob_{Y|\mathbf{X}}(-1|\mathbf{x})C_N\\
\mathrm{sign}\left[\hat{f}_0\left(\mathbf{x}\right)-\phi\right]=\mathrm{sign}\left[f_0^*(\mathbf{x})-\phi\right] & \text{if } \Prob_{Y|\mathbf{X}}(1|\mathbf{x})C_P \neq \Prob_{Y|\mathbf{X}}(-1|\mathbf{x})C_N
\end{cases}
\end{array}
}
\end{equation}
\vspace{4pt}

Bearing in mind that AdaBoost predictor $\hat{f}_0(\mathbf{x})$ is geared to satisfy (\refeq{costinsens_cond}), the optimality conditions for \emph {threshold modification} are not necessarily fulfilled. The only way to meet these requirements for any cost would be that the predictor obtained by AdaBoost matched the optimal one along the whole space, which is an obviously stronger condition than actually required (\refeq{costsens_cond}). Moreover, recalling the exponential bounding equation in which AdaBoost is based (\refeq{bound_ineq_eqn}), we can see that, once the sign of the obtained predictor matches the right label, the error bound is further minimized for increasing absolute values of the estimated predictor, no matter how close they are (or not) to the optimal predictor value.

As a consequence, there are no guarantees that a threshold change on the classical AdaBoost predictor will give us a cost-sensitive detector oriented to be optimal. Nonetheless, this non-optimality has not prevented that asymmetric detectors obtained by the Viola-Jones framework have been very successfully used for object detection.

\subsubsection{Heuristic}
\label{subsec:Heuristic} 

Most of the proposed cost-sensitive variations of AdaBoost \citep{Fan99, Ting00, ViolaJones02, Sun07} try to deal with asymmetry through direct manipulations of the weight update rule (\refeq{weight_rule_eqn}), but they are not full reformulations of AdaBoost for cost-sensitive scenarios. Masnadi-Shirazi and Vasconcelos pointed out that this kind of manipulations ``provide no guarantees of asymptotic convergence to a good cost-sensitive decision rule'' \citep{MasnadiVasconcelos11}, considering those algorithms as ``heuristic'' modifications of AdaBoost \citep{MasnadiVasconcelos07, MasnadiVasconcelos11}. 

Although these proposals have, in greater or lesser extent, some theoretical basis, for the sake of clarity and distinctiveness in our analysis, we will maintain the term \emph{heuristic}, as used in \citep{MasnadiVasconcelos07, MasnadiVasconcelos11}, to label this group of approaches based on the arbitrary modification of the weight update rule, as opposed to the full \emph{theoretical} derivations we will delve into in the next subsection.

\subsubsection*{AsymBoost}
\label{subsubsec:AsymBoost} 

Assuming the non-optimality of the strong classifier threshold adjustment procedure in their object detector framework (Section \ref{subsec:APosteriori}), Paul Viola and Michael J. Jones proposed a different scheme, coined as AsymBoost \citep{ViolaJones02}, trying to optimize AdaBoost for cost-sensitive classification problems.

Discarding the asymmetric weight initialization to be ``naive'' and only ``somewhat effective'' due to ``AdaBoost's balanced reweighting scheme'' (we will discuss on this point in Section \ref{subsec:weight}), AsymBoost proposes to distributedly emphasize weights by an asymmetric modulation before each round. In practical terms, the only change is multiplying weights $D(i)$ by a constant factor $(C_P/C_N)^{y_i/2T}$ before every learning step of a $T$-round process. As a consequence, the overall asymmetric factor seen by positive elements across the whole process is $C_P/C_N$ times the factor seen by negatives. 

\begin{equation}
\label{asb_equation}
D(i)_{t+1} = \frac{D_{t}(i)\exp\left(-\alpha_{t}y_{i}h_{t}\left(\mathbf{x}_i\right)\right)\left(\frac{C_P}{C_N}\right)^{\frac{y_{i}}{2T}}}{\sum_{i=1}^{n}D_{t}(i)\exp\left(-\alpha_{t}y_{i}h_{t}\left(\mathbf{x}_i\right)\right)\left(\frac{C_P}{C_N}\right)^{\frac{y_{i}}{2T}}}
\end{equation}

AsymBoost, that reduces to AdaBoost when costs are uniform, is detailed in Algorithm \ref{asb_algorithm} (Appendix \ref{app:algorithms}).

Though the global AsymBoost procedure seems to be theoretically sound, the \emph{equitable} asymmetry sharing among a \emph{fixed} number of rounds entails significant problems: Why such a rigid equitable sharing procedure should be optimal inserted in an adaptive framework such as AdaBoost? Why should we have to know in advance the number of training rounds while standard AdaBoost does not require that? Note that standard AdaBoost allows flexible performance tests to decide when to stop training, since any change in the total number of rounds is directly performed by training new additional rounds or trimming the current ensemble. However, a change in the size of the final ensemble (number of rounds) would strictly require Asymboost to re-train the whole classifier with a new asymmetry distribution.

\subsubsection*{AdaCost}

Wei Fan et al. proposed \citep{Fan99} a cost-sensitive variation of AdaBoost called AdaCost. The idea behind AdaCost is to modify the weight update rule, so examples with higher costs have sharper increases of their weights after misclassification but lighter decreases when are succesfully classified. This scheme is essentially addresed by introducing a misclassification adjustment function $\beta(i)$ into the weight update rule (\refeq{adc_weights}). 

\begin{equation}
\label{adc_weights}
D_{t+1}(i) = \frac{D_{t}(i)\exp\left(-\alpha_{t}y_{i}h_{t}\left(\mathbf{x}_i\right)\beta(i)\right)}{\sum_{i=1}^{n}D_{t}(i)\exp\left(-\alpha_{t}y_{i}h_{t}\left(\mathbf{x}_i\right)\beta(i)\right)}
\end{equation}

The misclassification adjustment function must depend on the cost ($C(i)$) of each example/class and the success/fail of its classification. As a result, $\beta(i)$ is imposed to be non-decreasing respect to $C(i)$ when classification fails, and non-increasing when classification succeeds. This opens the door to a huge amount of functions satisfying such requirements, from which authors chose the next:

\begin{equation}
\beta(i)=
\left\{
\begin{array}{ll}
0.5 \left(1-C(i)\right) & \mbox{$\text{if } h_{f}(\mathbf{x}_{i}) = y_{i}$},\\
0.5 \left(1+C(i)\right) & \mbox{$\text{if } h_{f}(\mathbf{x}_{i}) \neq y_{i}$}.
\end{array} \right.
\end{equation}

As can be seen, AdaCost does not match with AdaBoost for uniform costs and also applies a cost-dependent weight pre-emphasis (see Algorithm \ref{adc_algorithm}).

\subsubsection*{CSB0, CSB1 and CSB2}

Following the same idea of modifying the weight update rule, the CSB (acronym from Cost-Sensitive Boosting) family of algorithms \citep{Ting98,Ting00} propose three different updating schemes depending on which parameters are involved, resulting in CSB0, CSB1 and CSB2 algorithms (see respective Algorithms \ref{csb0_algorithm}, \ref{csb1_algorithm} and \ref{csb2_algorithm}). These rules are complemented, for all the three alternatives, by an asymmetric weight initialization and a minimum expected cost criterion for strong classification replacing the usual weighted voting scheme:

\begin{equation}
H(\mathbf{x})=\mathrm{sign}\left(\sum_{t=1}^{T}\alpha_{t}h_{t}(\mathbf{x}) \left( C_P \llbracket h_{t}(\mathbf{x})=+1 \rrbracket + C_N \llbracket h_{t}(\mathbf{x})=-1 \rrbracket \right)\right)
\end{equation}

This new voting rule gives emphasis, in run time, to weak hypothesis deciding in favor of the costly class. Of the three alternatives, only the last one, CSB2, is reduced to standard AdaBoost when costs are equal.

\subsubsection*{AdaC1, AdaC2 and AdaC3}

Defining new ways to modify the weight update rule, Yanmin Sun et al. \citep{Sun05, Sun07}  proposed another family of asymmetric AdaBoost alternatives called AdaC1, AdaC2 and AdaC3. These variants couple the cost factor in different parts of the update equation: inside the exponent (AdaC1), outside the exponent (AdaC2) and both (AdaC3):

\begin{equation}
\label{ac1_update}
D_{t+1}(i) = \frac{D_{t}(i)\exp\left(-\alpha_{t}c_{i}y_{i}h_{t}\left(\mathbf{x}_i\right)\right)}{\sum_{i=1}^{n}D_{t}(i)\exp\left(-\alpha_{t}c_{i}y_{i}h_{t}\left(\mathbf{x}_i\right)\right)}
\end{equation}

\begin{equation}
\label{ac2_update}
D_{t+1}(i) = \frac{c_{i}D_{t}(i)\exp\left(-\alpha_{t}y_{i}h_{t}\left(\mathbf{x}_i\right)\right)}{\sum_{i=1}^{n}c_{i}D_{t}(i)\exp\left(-\alpha_{t}y_{i}h_{t}\left(\mathbf{x}_i\right)\right)}
\end{equation}

\begin{equation}
\label{ac3_update}
D_{t+1}(i) = \frac{c_{i}D_{t}(i)\exp\left(-\alpha_{t}c_{i}y_{i}h_{t}\left(\mathbf{x}_i\right)\right)}{\sum_{i=1}^{n}c_{i}D_{t}(i)\exp\left(-\alpha_{t}c_{i}y_{i}h_{t}\left(\mathbf{x}_i\right)\right)}
\end{equation}

As a difference from previous approaches, these changes in the weight update are also propagated to the way goodness parameter $\alpha_t$ is defined and, as a consequence, have influence on how the weak classifier error is computed (see Algorithms \ref{ac1_algorithm}, \ref{ac2_algorithm}, \ref{ac3_algorithm}). All these variants reduce to AdaBoost when the cost function $C(i)$ is 1 for all examples.

\subsubsection{Theoretical}
\label{subsec:Theoretical}

The methods in the previous subsection have one key point in common: the starting point of their derivations is an arbitrary modification of the weight update rule. However, as can be easily shown following the work by Schapire and Singer \citep{SchapireSinger99}, weight update in standard AdaBoost is actually a \emph{consequence} of the error minimization procedure (\refeq{bound_ineq_eqn}) and not an arbitrary starting point of it. Thus, the way to reach theoretically sound cost-sensitive boosting algorithms should be to walk the path in the opposite direction: designing a new asymmetric derivation scheme to obtain a new full formulation (that may include a new weight update rule), instead of partially adapting previous equations.

There are three alternatives in the literature that follow different theoretically sound derivation schemes reaching cost-sensitive variants of AdaBoost: Cost-Sensitive AdaBoost \citep{MasnadiVasconcelos07, MasnadiVasconcelos11}, AdaBoostDB \citep{LandesaAlba13} and Cost-Generalized AdaBoost \citep{LandesaAlba12}.

\subsubsection*{Cost-Sensitive AdaBoost}

The Cost-Sensitive Boosting framework proposed by Hamed Masnadi-Shirazi and Nuno Vasconcelos \citep{MasnadiVasconcelos07, MasnadiVasconcelos11} has its roots in the Statistical View of Boosting \citep{Friedman00}, by adapting the standard loss in equation (\refeq{loss_eqn}) with asymmetric exponential arguments for each class component.

\begin{equation} 
\label{csloss_eqn}
J(f(\mathbf{x}))=\E\left[\llbracket y=1 \rrbracket \mathrm{e}^{-C_{P}f(\mathbf{x_i})} + \llbracket y=-1 \rrbracket \mathrm{e}^{C_{N}f(\mathbf{x_i})}\right]
\end{equation}

This asymmetric loss is theoretically minimized by the asymmetric logistic transform of $\Prob\left(y=1|\mathbf{x}\right)$ (see Section \ref{subsec:Statistical View}), which should ensure cost-sensitive optimality.

\begin{equation}
\label{stat_sol_asym}
\begin{split}
f(x)=\frac{1}{C_{P}+C_{N}}\log\frac{C_{P}\Prob\left(y=1|\mathbf{x}\right)}{C_{N}\Prob\left(y=-1|\mathbf{x}\right)}
\end{split}
\end{equation}

The empirical minimization of the asymmetric loss proposed by Masnadi-Shirazi and Vasconcelos follows a gradient descent scheme on the space of boosted (combined and modulated) binary weak classifiers, resulting in the Cost-Sensitive Adaboost algorithm shown in Algorithm \ref{csa_algorithm}. As can be seen, the final solution involves hyperbolic functions and scalar search procedures, being extremely more complex and computing demanding than the original AdaBoost.

\subsubsection*{AdaBoostDB}

Following the generalizad analysis of AdaBoost \citep{SchapireSinger99} instead of the Statistical View of Boosting, a different approach to provide AdaBoost with Cost-Sensitive properties through a fully theoretical derivation procedure is presented in \citep{LandesaAlba13}. This algorithm, coined as AdaBoostDB (from Double Base), is based on the use of different exponential bases $\beta_P$ and $\beta_N$ for each class error component, thus defining a class-dependent error bound to minimize.

\begin{equation}
\label{exp_bound_eqn_asym}
E_{T} \leq \tilde{E}_T = \sum_{i=1}^{m} D_{1}(i){\beta_P}^{ -y_{i} f(\mathbf{x_{i}})} + \sum_{i=m+1}^{n} D_{1}(i){\beta_N}^{ -y_{i} f(\mathbf{x_{i}})}
\end{equation}

On the one hand, the derivation scheme followed and the polynomial model used to address the problem, enable a different and extremely efficient formulation, able to achieve over 99\% training time saving with respect to Cost-Sensitive AdaBoost (see Algorithm \ref{abdb_algorithm}). On the other hand, this class-dependent error is fully equivalent to the cost-sensitive loss (\refeq{csloss_eqn}) defined for Cost-Sensitive Boosting, so both minimizations will converge to the same solution and ensure the same formal guarantees. 

As a result, AdaBoostDB is a much more efficient framework to reach the same solution as Cost-Sensitive Boosting (except for numerical errors related to the different models adopted, hyperbolyc vs. polynomial). However, despite its large improvement in training complexity and performance, AdaBoostDB is still much more complex than standard AdaBoost.

\subsubsection*{Cost-Generalized AdaBoost}

The Asymmetric AdaBoost problem is addressed in \citep{LandesaAlba12} from a different theoretical perspective, realizing that one kind of modification have systematically been either overlooked or undervalued in the related literature: weight initialization. 

Even though some preliminary studies by Freund and Schapire \citep{FreundSchapire97}, creators of AdaBoost, left the initial weight distribution free to be controlled by the learner, AdaBoost is ``de facto'' defined, almost everywhere in the literature (e.g. \citep{Schapire98, SchapireSinger99, Fan99, FreundSchapire99, Ting00, Friedman00, Polikar06, Sun07, Polikar07, MasnadiVasconcelos11}), with a fixed initial uniform weight distribution. From there, some asymmetric boosting algorithms (like AdaCost or CSB) use cost-sensitive initialization as a lateral or secondary strategy respect to their proposed weight update rules, while others (like AsymBoost or Cost-Sensitive Boosting), immediately discard asymmetric weight initialization to be ``naive'' and ineffective, arguing that the first boosting round would absorb the full introduced asymmetry and the rest of the process would keep entirely symmetric. 

In \citep{LandesaAlba12}, following a different insight to analyze AdaBoost and obtaining a novel error bound interpretation, asymmetric weight initialization is shown to be an effective way to reach cost-sensitiveness, and, as occurs with everything related to boosting, it is achieved in an additive round-by-round (asymptotic) way. All, with the added advantage that weight initialization is the only needed change to gain asymmetry with regard to standard AdaBoost (even weight update rule is unchanged). Hence, for whatever desired asymmetry, both complexity and formal guarantees of the original AdaBoost remain intact. 

In this work, we will refer to the algorithm underlying this perspective as Cost-Generalized AdaBoost (see Algorithm \ref{adbg_algorithm}).

\section{Theoretical Algorithms: Analysis and Discussion}
\label{sec:Discuss}

Though in the experimental part of our work (see the accompanying paper \citep{LandesaAlba??b}) we will show comparative results of all the alternatives presented in the previous section, at this point we will focus our attention on the three proposals with a fully theoretical derivation scheme: Cost-Sensitive AdaBoost \citep{MasnadiVasconcelos07,MasnadiVasconcelos11}, AdaBoostDB \citep{LandesaAlba13} and Cost-Generalized AdaBoost \citep{LandesaAlba12}. The first important aspect we should notice is that these three proposals can be effectively analyzed as if they were only two, since Cost-Sensitive AdaBoost and AdaBoostDB, despite following different perspectives and obtaining markedly different algorithms, share an equivalent theoretical root and drive to the same solution \citep{LandesaAlba13}. As a consequence, if not otherwise specified, in this section we will refer to one or another interchangeably, giving priority to the name Cost-Sensitive AdaBoost due to its chronological precedence.

\subsection{The Question of Weight Initialization}
\label{subsec:weight}

As commented in Section \ref{subsec:Theoretical}, despite some initial studies pointing to free initial weight distributions \citep{FreundSchapire97} or works proposing cost-proportional weighting as an effective way to transform generic cost-insensitive learning algorithms into cost-sensitive ones \citep{Zadrozny03}, subsequent works on boosting have insisted on two recurrent ideas: On the one hand, uniform distribution has been assumed as the ``de facto'' standard  for weight initialization when defining AdaBoost (e.g. \citep{Schapire98, SchapireSinger99, Fan99, FreundSchapire99, Ting00, Friedman00, Polikar06, Sun07,Polikar07, MasnadiVasconcelos11}); on the other hand, asymmetric weight initialization has been systematically rejected as a valid method to achieve cost-sensitive boosted classifiers, arguing that it is insufficient \citep{Fan99, Ting00} or ineffective \citep{ViolaJones02, MasnadiVasconcelos07, MasnadiVasconcelos11}. 

However, in \citep{LandesaAlba12}, AdaBoost is demonstrated to have inherent and sound cost-sensitive properties embedded in the way the weight distribution is initialized. In fact, the method we are referring to as Cost-Generalized AdaBoost, was not even originally proposed as a new algorithm: ``it is just AdaBoost'' \citep{LandesaAlba12} with appropriate initial weights. Such an analysis, supported by a novel class-conditional interpretation of AdaBoost, is, thus, in clear contradiction to the supposed ineffectiveness of cost-sensitive weight initialization underlying previous works.

In order to definitely clarify this contradiction, we will connect both perspectives by demonstrating the validity of asymmetric weight initialization in the same scenarios and lines of reasoning that have been previously used in the literature to decline its use.

\subsubsection{The Supposed Symmetry}
\label{subsubsec:supposed_symmetry}

Masnadi-Shirazi and Vasconcelos \citep{MasnadiVasconcelos11} when explaining their Cost-Sensitive Boosting framework, immediately discard the unbalanced weight initialization (calling it ``naive implementation'') with the argument that iterative weight update in AdaBoost ``quickly destroys the initial asymmetry'' obtaining a ``predictor'' which ``is usually not different from that produced with symmetric initial conditions''. Though their statement is not explicitly supported for any further test or bibliographic reference, it seems to be extracted from the work by Viola and Jones \citep{ViolaJones02} in which AsymBoost is presented. In that work, the initial weight modification technique is rejected arguing that ``the first classifier selected absorbs the entire effect of the initial asymmetric weights'', and assuming the rest of the process as ``entirely symmetric''. It is because of this seeming problem that AsymBoost was designed for distributing an equitable asymmetry among a fixed number of rounds.

\begin{figure}[!htb]
\centering
\includegraphics[width=0.8\columnwidth]{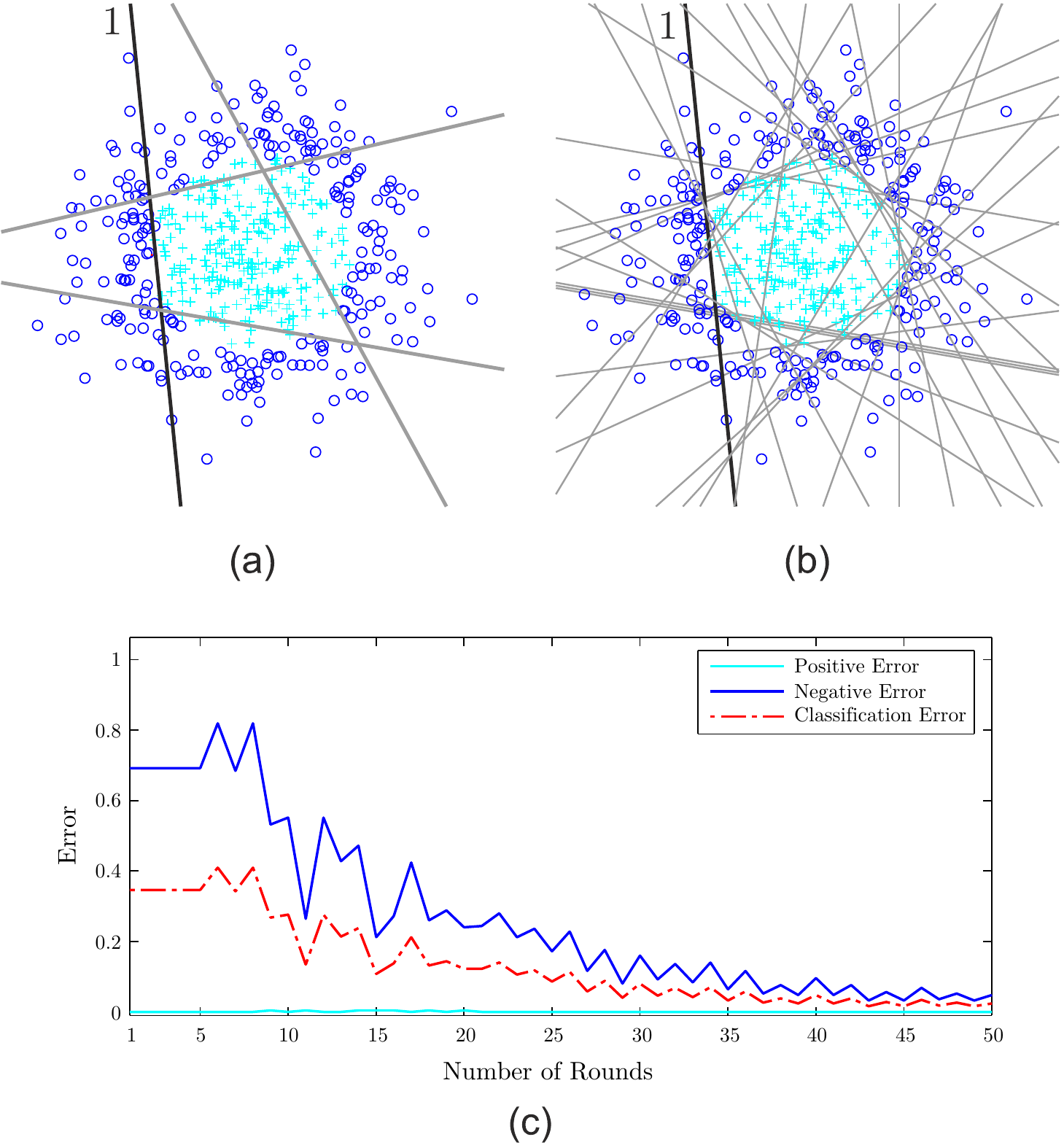}
\caption{Synthetic counterexample to the example by Viola and Jones \citep{ViolaJones02}, with costs $C_P=4$ and $C_N=1$, and the same polarity as the original:(a) training set with the first four weak classifiers superimposed; (b) weak classifiers after 50 training rounds; (c) Global error evolution through 50 training rounds. Weak classifiers are stumps in the linear 2D space. Positive examples are marked as `$+$', `$\circ$' are the negative ones, and `1' denotes the first selected weak classifier. Positives are the costly class.}
\label{counter_example_1_fig} 
\end{figure}

The cost-sensitive analysis by Viola and Jones \citep{ViolaJones02} is illustrated by a four-round boosted classifier graphic representation that supports their conclusions against asymmetric weight initialization. However, this example can be misleading: what would happen if boosting were run for more than those four rounds? An answer can be found in Figure \ref{counter_example_1_fig}, where we have reproduced and extended that illustrative experiment.

Strictly following Viola and Jones \citep{ViolaJones02}, after Figure \ref{counter_example_1_fig}a we could reach the seeming conclusion that, once an initial asymmetric weak classifier has been selected, the selection of the remaining weak classifiers is not guided by an asymmetric goal. However, as showed by Schapire and Singer \citep{SchapireSinger99}, AdaBoost is an additive minimization process and, as such, it has an \emph{asymptotic} behavior, a kind of behavior that can not be properly judged by stopping after only a few training rounds. Running the algorithm for many more rounds in the same example (see Figure \ref{counter_example_1_fig}b), we appreciate that many other subsequent selected classifiers are, at least, as asymmetric as the first one.

The class-conditional interpretation of AdaBoost in \citep{LandesaAlba12} shows that the asymmetry encoded by the initial weight distribution is actually translated to a cost-sensitive global error (a weighted error), and what AdaBoost is actually minimizing is a bound on that global error. Thus, instead of inspecting the individual asymmetry of each single hypothesis, the cost-sensitive behavior of AdaBoost should be evaluated, for correctness, in terms of the \emph {cumulative contribution} of \emph{all} the selected weak classifiers giving rise to the strong one. Figure \ref{counter_example_1_fig}c shows how, even in a scenario like the one proposed by Viola and Jones \citep{ViolaJones02}, the classifier obtained by AdaBoost after an asymmetric weight initialization follows a real cost-sensitive iterative profile. 

\begin{figure}[!htb]
\centering
\includegraphics[width=0.8\columnwidth]{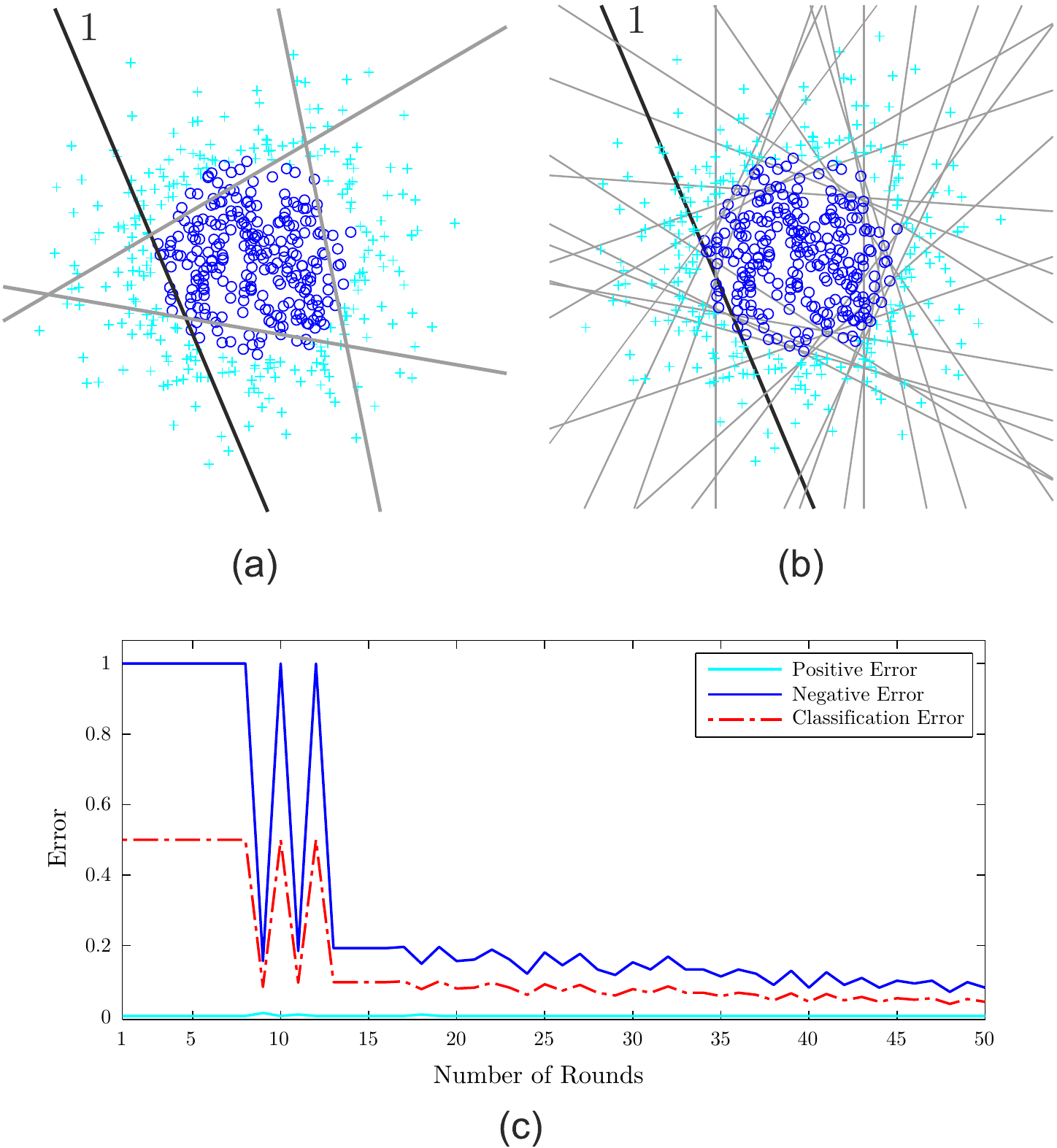}
\caption{Synthetic counterexample to the example by Viola and Jones \citep{ViolaJones02}, with costs $C_P=4$ and $C_N=1$, and with opposite polarity to the original:(a) training set with the first four weak classifiers superimposed;(b) weak classifiers after 50 training rounds; (c) Global error evolution through 50 training rounds. Weak classifiers are stumps in the linear 2D space. Positive examples are marked as `$+$', `$\circ$' are the negative ones, and `1' denotes the first selected weak classifier. Positives are the costly class.}
\label{counter_example_2_fig} 
\end{figure}

Moreover, postulates by Viola and Jones \citep{ViolaJones02} and Masnadi-Shirazi and Vasconcelos \citep{MasnadiVasconcelos11} can also be refuted by simply inverting labels on the same set (see Figure \ref{counter_example_2_fig}). As can be seen, no weak classifier is able to satisfy, by itself, the requirements of that ``supposed'' initial round absorbing the full asymmetry of the problem. However, even in such an unfavorable scenario, the desired asymmetry is effectively achieved, from cost-proportionate initial weights, after a (boosted) round-by-round cumulative process.

Further comments on these experiments can be found in Appendix \ref{app:comments_figures}.

\subsubsection{Weight Initialization inside the Cost-Sensitive Boosting Framework}
\label{subsubsec:weight_cost_sensitive}

Cost-Sensitive AdaBoost \citep{MasnadiVasconcelos11} is an algorithm that, despite having a rigorous theoretical derivation, is built upon the belief that cost-sensitive initial weighting is not a valid method to achieve asymmetric boosted classifiers. However, as we have already mentioned, the theoretical analysis in \citep{LandesaAlba12} refutes that supposed invalidity. A clarifying experiment at this point is to introduce asymmetric weight initialization inside the Cost-Sensitive AdaBoost theoretical framework, to assess the theoretical validity of the former with the tools used by the latter.

Based on the Statistical View of Boosting \citep{Friedman00}, the cost-sensitive expected loss (i.e. the  risk function) proposed by Masnadi-Shirazi and Vasconcelos to derive Cost-Sensitive AdaBoost, consists on two class-dependent exponential components with asymmetry embedded in its exponents:

\begin{equation}
\label{csloss_eqn1}
J_{CSA}(f(\mathbf{x}))=\E\left[\llbracket y=1 \rrbracket \exp(-C_{P}f(\mathbf{x})) + \llbracket y=-1 \rrbracket \exp(C_{N}f(\mathbf{x}))\right]
\end{equation}

Following the proof derivation scheme in \citep{MasnadiVasconcelos11}, if the derivatives of this loss are set to zero, we will obtain the function of minimum expected loss (minimum risk) conditioned on $\mathbf{x}$ for Cost-Sensitive AdaBoost, that, as can be seen, is based on the asymmetric logistic transform of $\Prob(y=1|\mathbf{x})$.

\begin{equation}
\label{statistical_deriv_cda}
{\begin{array}{c} 
J_{CSA}(f(\mathbf{x}))=\Prob\left(y=1|\mathbf{x}\right)\exp(-C_{P}f(\mathbf{x}))+\Prob\left(y=-1|\mathbf{x}\right)\exp(C_{N}f(\mathbf{x}))\\
\Downarrow\\
\dfrac{\partial J_{CSA}(f(\mathbf{x}))}{\partial f(\mathbf{x})}= -C_{P}\Prob\left(y=1|\mathbf{x}\right)\exp(-C_{P}f(\mathbf{x}))+C_{N}\Prob\left(y=-1|\mathbf{x}\right)\exp(C_{N}f(\mathbf{x}))=0\\
\Downarrow\\
\dfrac{C_{P}\Prob\left(y=1|\mathbf{x}\right)}{C_{N}\Prob\left(y=-1|\mathbf{x}\right)}=\exp((C_{P}+C_{N})f(\mathbf{x}))\\
\Downarrow\\
f_{CGA}(\mathbf{x})=\dfrac{1}{C_{P}+C_{N}}\log\left(\dfrac{C_{P}\Prob\left(y=1|\mathbf{x}\right)}{C_{N}\Prob\left(y=-1|\mathbf{x}\right)}\right)
\end{array}
}
\end{equation}

Now, let us suppose that the two cost parameters $C_P$ and $C_N$, rather than in the exponents, are incorporated as direct modulators of the exponentials (\refeq{csloss_exp}). This procedure is equivalent to model the initial weight distribution by means of two uniform \emph{class-conditional} distributions, respectively modulated by $C_{P}/\left(C_{P}+C_{N}\right)$ and $C_{N}/\left(C_{P}+C_{N}\right)$, i.e. an asymmetric weight initialization as the one proposed giving rise to Cost-Generalized AdaBoost.

\begin{equation}
\label{csloss_exp}
J_{CGA}(f(\mathbf{x}))=\E\left[\llbracket y=1 \rrbracket C_{P} \exp(-f(\mathbf{x})) + \llbracket y=-1 \rrbracket C_{N} \exp(f(\mathbf{x}))\right]
\end{equation}

If we repeat the above derivation scheme on this new loss, we will find the function of minimum expected loss (minimum risk) conditioned on $\mathbf{x}$ for Cost-Generalized AdaBoost:

%
%
%

\begin{equation}
\label{statistical_deriv_cga}
{\begin{array}{c} 
J_{CGA}(f(\mathbf{x}))=\Prob\left(y=1|\mathbf{x}\right)C_{P}\exp(-f(\mathbf{x}))+\Prob\left(y=-1|\mathbf{x}\right)C_{N}\exp(f(\mathbf{x}))\\
\Downarrow\\
\dfrac{\partial J_{CGA}(f(\mathbf{x}))}{\partial f(\mathbf{x})}= -\Prob\left(y=1|\mathbf{x}\right)C_{P}\exp(-f(\mathbf{x}))+\Prob\left(y=-1|\mathbf{x}\right)C_{N}\exp(f(\mathbf{x}))=0\\
\Downarrow\\
\dfrac{C_{P}\Prob\left(y=1|\mathbf{x}\right)}{C_{N}\Prob\left(y=-1|\mathbf{x}\right)}=\exp(2f(\mathbf{x}))\\
\Downarrow\\
f_{CGA}(\mathbf{x})=\dfrac{1}{2}\log\left(\dfrac{C_{P}\Prob\left(y=1|\mathbf{x}\right)}{C_{N}\Prob\left(y=-1|\mathbf{x}\right)}\right)
\end{array}
}
\end{equation}

%
%
%

As can be seen, the obtained minimizer is also based on the asymmetric logistic transform of $\Prob(y=1|\mathbf{x})$, showing us that, even from the Cost-Sensitive AdaBoost derivation perspective, there is no reason to discard asymmetric weight initialization as a valid approach to build cost-sensitive boosted classifiers\footnote{As analyzed in Appendix \ref{app:weight}, the way asymmetry is applied across the different boosting variants covered by the Cost-Sensitive Boosting framework \citep{MasnadiVasconcelos11} is not homogeneus either. In fact, despite having discarded cost-proportionate weight initialization as a valid method, one of the algorithms (Cost-Sensitive LogitBoost) proposed in the same work is actually based on that strategy.}.

\subsection{Comparative Analysis of the Theoretical Approaches}
\label{subsec:algorithms_cmp}

As we have seen, among the three asymmetric AdaBoost algorithms with a full theoretical derivation, two of them (Cost-Sensitive AdaBoost and AdaBoostDB) drive to the same solution, while the other one (Cost-Generalized AdaBoost) has been shown to guarantee the same theoretical validity than its counterparts. At this point, we may wonder if Cost-Generalized AdaBoost is also obtaining the same solution as Cost-Sensitive AdaBoost/AdaBoostDB. As we will see in the experimental part of our work (in the second part of this series of two papers \citep{LandesaAlba??b}) the answer to this question is ``no'': classifiers obtained by Cost-Sensitive AdaBoost and Cost-Generalized AdaBoost in the same scenarios are markedly different. In this section, from a theoretical perspective, we will analyze the differences between the two algorithms, with the aim of achieving the intrinsic distinctivenesses of their respective classifiers.

\subsubsection{Error Bound Minimization}
\label{subsubsec:error_bound_cmp}

As commented in Section \ref{sec:CSvar}, the most common detection problem can be parametrized by the next cost matrix:

\begin{equation}
\mathbf{C}
=\left(
\begin{array} {c c}
c_{nn} & c_{np}\\
c_{pn} & c_{pp}\\
\end{array}
\right)
=\left(
\begin{array} {c c}
0 & C_P\\
C_N & 0\\
\end{array}
\right)
\end{equation}

\begin{figure}[!htb]
\centering
\includegraphics[width=7.5cm]{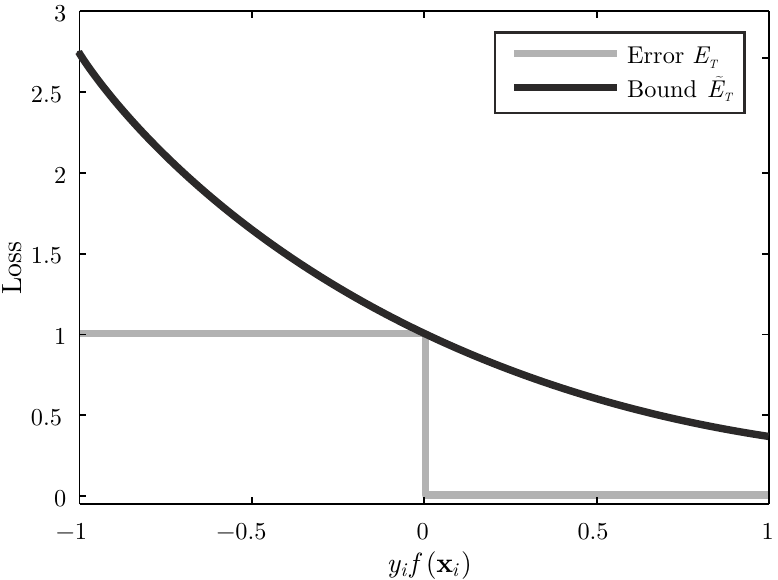}
\caption{Training error bound of AdaBoost. The loss (y-axis) associated to each decision has an exponential dependency on the performance score of the strong classifier (x-axis).}
\label{std_adb_bound_fig} 
\end{figure}

We will start our comparative analysis by following the error bound minimization perspective originally proposed by Schapire and Singer \citep{SchapireSinger99}, also used in the derivation of Cost-Generalized AdaBoost and AdaBoostDB. From that point of view, classical AdaBoost, with its initial uniform weight distribution, is an algorithm driven to minimize an exponential bound ($\tilde{E}_T$) on the training error ($E_T$) (\refeq{std_adb_bound}), as illustrated in Figure \ref{std_adb_bound_fig}. In that figure, the horizontal axis ($y_if(\mathbf{x}_i)$) represents the \emph{performance score} of a classification, whose sign indicates the success (if $y_if(\mathbf{x}_i)>0$) or failure (if $y_if(\mathbf{x}_i)<0$) of the decision, and whose magnitude indicates the confidence expected by the classifier on its decision. The exponential bound is decreasing for increasing performance scores, so the classical AdaBoost minimization process is aimed to maximize correct classifications and their margin (distance to the boundary), in a scenario where all the training examples follow a common cost scheme.
 
\begin{equation}
\label{std_adb_bound}
E_{T}= \sum_{i=1}^{n} \frac{1}{n} \llbracket H(\mathbf{x}_{i}) \neq y_{i}\rrbracket \leq \sum_{i=1}^{n} \frac{1}{n} \exp \left( -y_{i} f(\mathbf{x}_{i}) \right) = \tilde{E}_{T}
\end{equation}

\begin{figure}[!htb]
\centering
\includegraphics[width=7.5cm]{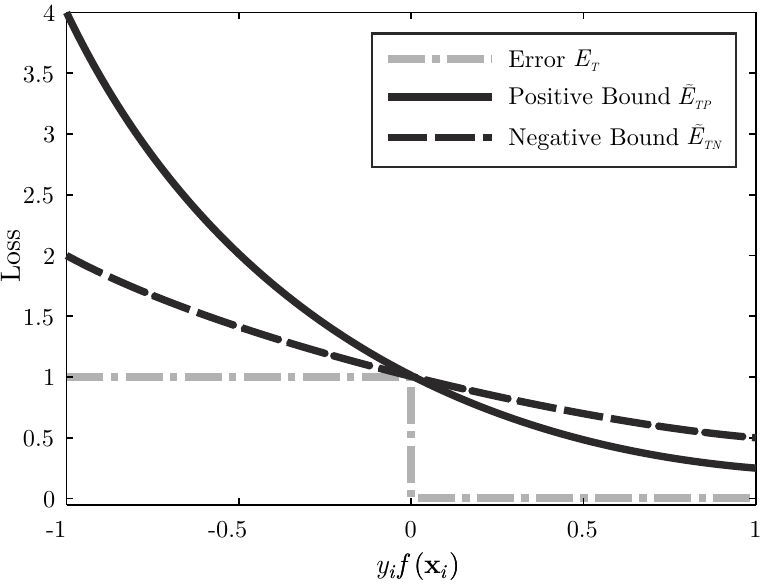}
\caption{Training error bound of Cost-Sensitive AdaBoost and AdaBoostDB for $C_P=2$ and $C_N=1$. Loss has a class-dependent definition and is composed of two different exponential functions.}
\label{csb_bound_fig} 
\end{figure}

Cost-Sensitive AdaBoost and AdaBoostDB, assumming that the training set is divided into two significant subsets (positives and negatives), define two different exponential bounds ($\tilde{E}_{TP}$ and $\tilde{E}_{TN}$) with different associated costs ($C_P$ and $C_N$) over each subset. These costs are inserted as exponent modulators into each class-dependent exponential bound (\refeq{csb_bound}), reaching a cost-sensitive behavior that can be graphically interpreted as shown in Figure \ref{csb_bound_fig}. The goal is, again, to maximize correct classifications and their margin, but this time in a scenario where positives and negatives have different associated losses.

\begin{equation}
\label{csb_bound}
\begin{split}
E_{T} &= \sum_{i=1}^{n} \frac{1}{n} \llbracket H(\mathbf{x}_{i}) \neq y_{i}\rrbracket\\
& \leq \sum_{i=1}^{m} \frac{1}{n} \exp \left( -C_{P}y_{i} f(\mathbf{x}_{i}) \right) + \sum_{i=m}^{n} \frac{1}{n} \exp \left( -C_{N}y_{i} f(\mathbf{x}_{i}) \right)\\
& =\tilde{E}_{TP} + \tilde{E}_{TN} = \tilde{E}_{T}
\end{split}
\end{equation}

As can be seen, asymmetric modifications in Cost-Sensitive AdaBoost (and AdaBoostDB) are based on new bounds for the training error, while the error definition itself remains unchanged from original (cost-insensitive) AdaBoost. 

\begin{figure}[!htb]
\centering
\includegraphics[width=7.5cm]{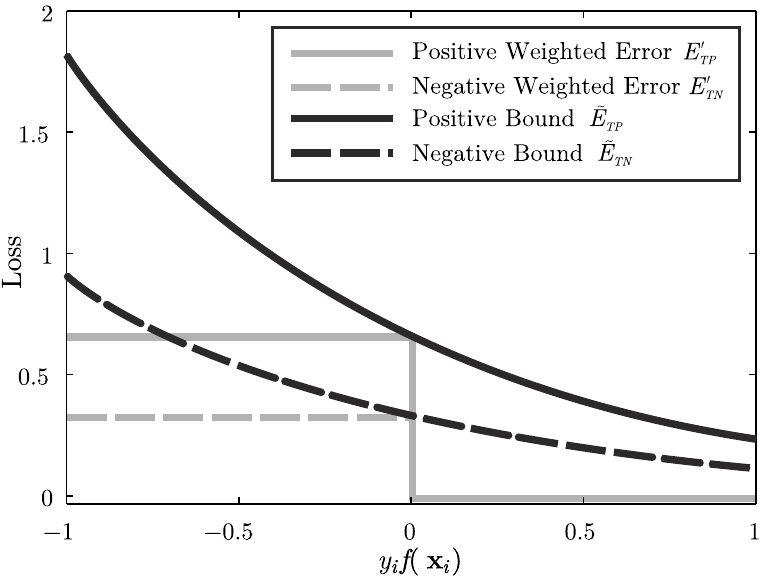}
\caption{Training error bound of Cost-Generalized AdaBoost for $C_P=2$ and $C_N=1$. Loss keeps again an exponential dependency, but now modulated by a class-dependent behavior.}
\label{cgb_bound_fig} 
\end{figure}

Cost-Generalized AdaBoost, on the other hand, is based on redefining the training error and then applying the standard exponential bounding process. To achieve this, training error in positives (${E}_{TP}$) and in negatives (${E}_{TN}$) are computed separately, and then are modulated by its respective normalized costs. The resulting class-dependent weighted error components (${E}_{TP}'$ and ${E}_{TN}'$) jointly define the \emph{cost-sensitive global training error} ($E_{T}'$). The same way as in standard AdaBoost, each of these weighted error components can be exponentially bounded ($\tilde{E}_{TP}$ and $\tilde{E}_{TN}$), and the combination of the two resulting \emph{class-dependent} bounds will define a cost-sensitive global bound ($\tilde{E}_{T}$) (\refeq{cgb_bound}), that is the function being minimized by Cost-Generalized AdaBoost. The scenario is graphically depicted in Figure \ref{cgb_bound_fig}.

\begin{equation}
\label{cgb_bound}
\begin{split}
E_{T}' &= E_{TP}' + E_{TN}' = \frac{C_{P}}{C_{P}+C_{N}} E_{TP} + \frac{C_{N}}{C_{P}+C_{N}} E_{TN} \\
& = \frac{C_{P}}{C_{P}+C_{N}} \sum_{i=1}^{m} \frac{1}{m} \llbracket H(\mathbf{x}_{i}) \neq y_{i}\rrbracket + \frac{C_{N}}{C_{P}+C_{N}} \sum_{i=m+1}^{n} \frac{1}{n-m} \llbracket H(\mathbf{x}_{i}) \neq y_{i}\rrbracket  \\
& \leq \frac{C_{P}}{C_{P}+C_{N}} \sum_{i=1}^{m} \frac{1}{m} \exp \left( -y_{i} f(\mathbf{x}_{i}) \right) + \frac{C_{N}}{C_{P}+C_{N}} \sum_{i=m}^{n} \frac{1}{n-m} \exp \left( -y_{i} f(\mathbf{x}_{i}) \right)\\
& =\tilde{E}_{TP} + \tilde{E}_{TN} = \tilde{E}_{T}
\end{split}
\end{equation}

It is important to notice that, by definition, all these algorithms have the goal of obtaining the best possible classifier able to deal with the problem in a cost-sensitive sense, and that the bounding loss functions $\tilde{E}_{T}$ are a mere mathematical tool to make the minimization problem tractable. Thus, from a formal point of view, the direct definition of a cost-sensitive error to be subsequently bounded, as proposed by Cost-Generalized AdaBoost, seems to be more suitable than using the standard cost-insensitive error and manipulate its bound to be asymmetric, as suggested by Cost-Sensitive AdaBoost or AdaBoostDB.

\begin{figure}[!htb]
\centering
\includegraphics[width=0.9\columnwidth]{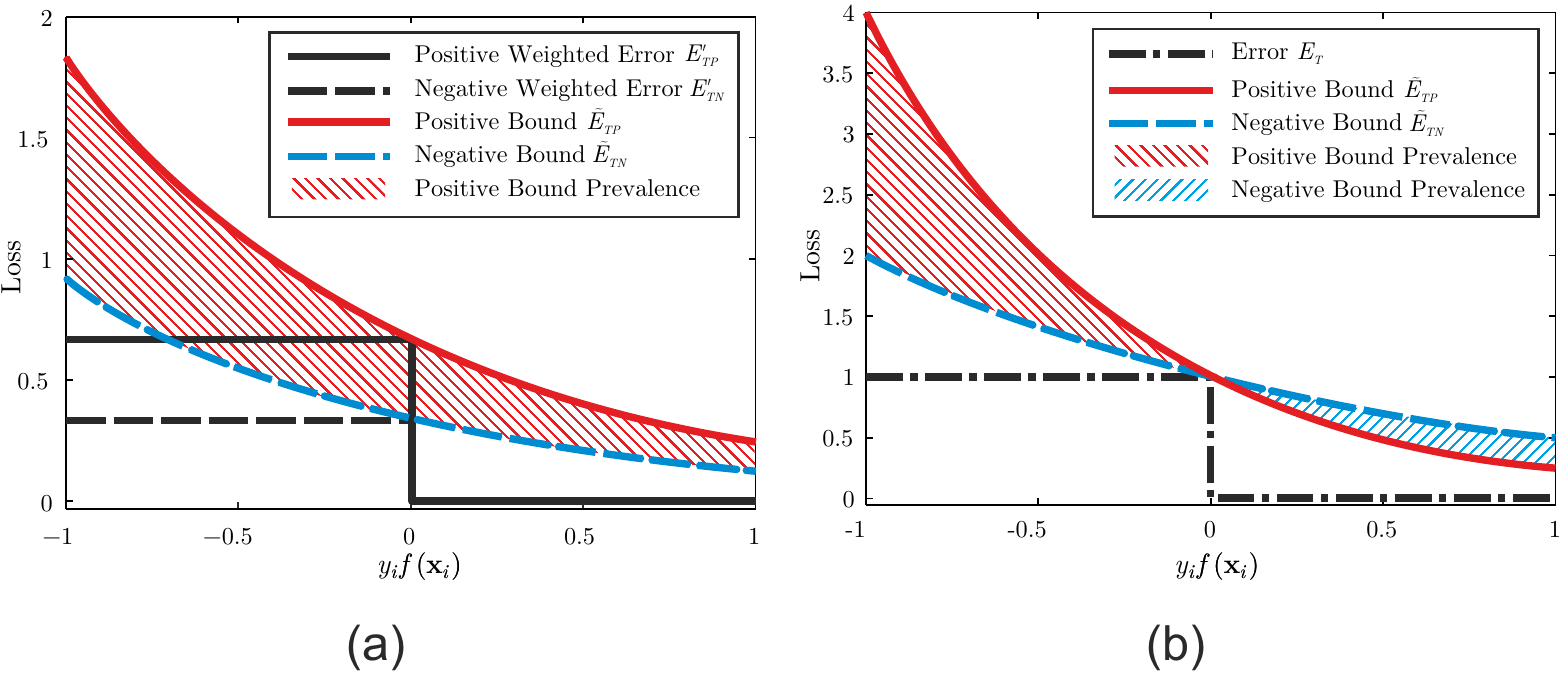}
\caption{Class prevalence of error bounds for Cost-Generalized AdaBoost (a) and Cost-Sensitive AdaBoost (b) ($C_P=2$, $C_N=1$).}
\label{preval_fig} 
\end{figure}

Figure \ref{preval_fig} illustrates the prevalence of the class-dependent error bounds of the two algorithms, assuming, without loss of generality, that positives have a greater cost than negatives $C_P>C_N$ (the opposite case can be modeled by a simple label swap). As can be seen, in Cost-Generalized AdaBoost (Figure \ref{preval_fig}a) the loss associated to positives is always greater than the loss associated to negatives, and the ratio between the two class-dependent losses remains constant along the performance scores. However, in Cost-Sensitive AdaBoost (Figure \ref{preval_fig}b), the ratio between losses varies according to the score, to the extent that class prevalence is inverted depending on which side of the success boundary ($y_if(\mathbf{x}_i)=0$) we are.

The iterative learning process behind AdaBoost builds a predictor function $f(\mathbf{x}_i)$ aimed to progressively (round by round) minimize the respective loss function over the training dataset. In terms of classification, this means that AdaBoost classifiers are trained not only to maximize the accuracy of the classifier over the training set, but also to maximize the margin of its decisions. So, once one training example is correctly classified, the tendency of the learner will be to continue increasing the confidence of its prediction ($\mathrm{abs}(f(\mathbf{x}_i))$) to move it away from the decision boundary ($f(\mathbf{x}_i)=0$). For Cost-Generalized AdaBoost, this means that any positive training example will always be more costly (and in the same ratio) than any negative example with its same performance score, whatever this score is. However, in the case of Cost-Sensitive AdaBoost, prevalence ratio varies exponentially with performance scores. So, when scores are positive, negative training examples become the prevalent ones. 

Bearing in mind that the performance score of any training example, at any iteration of the learning process, is determined by the evaluation over the example of the boosted predictor learned so far, and that the weight of this example for the next learning round will depend on the value of the related bounding loss for that particular score, we can draw the two following consequences:

\begin{itemize}
\item In Cost-Generalized AdaBoost positives will always be the costly class, and the same cost asymmetry is preserved throughout the whole learning process.
\item In Cost-Sensitive AdaBoost cost asymmetry changes. While the classifier is wrong, positives are the costly class (learning is positive-driven), but when classification is correct, negatives are prevalent (learning is negative-driven). The more accurate the classifier obtained is, the more costly will be negatives over positives in subsequent training rounds.
\end{itemize}

In terms of training error, these differences seem to be anecdotal, since the change of class prevalence occurs once the classifier succeeds for each example. However, what is really relevant, is the effect in terms of generalization error: when the classifier works on unseen instances it will make mistakes and it is essential, from a cost-sensitive perspective, to characterize which class is the most prone to errors and to what extent. 

As the iterative training process progresses, the performance scores associated to the training examples tend to increase, and their respective losses tend to decrease moving along the Y axis on Figures \ref{csb_bound_fig} and \ref{cgb_bound_fig}, so, the more rounds we train, the more on the right of these figures we will be. In the case of Cost-Sensitive AdaBoost this trend will increasingly emphasize negatives at the expense of positives, while Cost-Generalized AdaBoost keeps the ratio between classes intact throughout the whole learning process. Thus, due to its changing emphasis, Cost-Sensitive AdaBoost may run the risk of obtaining classifiers in which the supposed costly class is the most prone to errors: just the opposite of what was originally intended! 

In the companion paper of the series \citep{LandesaAlba??b} we will see empirical evidences confirming this \emph{asymmetry swapping} behavior that, by definition, is expected to be more noticeable the closer the system is to overfitting, but that may have an implicit detrimental effect on the performance reached by all classifiers trained by Cost-Sensitive AdaBoost.

\subsubsection{Statistical View of Boosting}
\label{subsubsec:statistical_cmp}

Instead of the exponential error bound minimization perspective that originally gave rise to AdaBoost (and that also is the derivation core of Cost-Generalized AdaBoost and AdaBoostDB) we will now adopt a different point of view: the Statistical View of Boosting \citep{Friedman00}, the other major analytical framework to interpret and derive AdaBoost that, in addition, is the foundation of Cost-Sensitive AdaBoost.

As we have seen in Section \ref{subsec:Statistical View}, from the Statistical View of Boosting perspective, AdaBoost can be interpreted as an algorithm that iteratively builds an additive regression model based on the following loss function:

\begin{equation}
\label{loss_ab_eqn}
l_{AB}(f(\mathbf{x}),y)=\exp\left(-yf(\mathbf{x})\right)
\end{equation}

From that loss, an associated risk function $J_{AB}(f(\mathbf{x}))$ (the expected loss) is defined:

\begin{equation}
\label{risk_ab_eqn}
\begin{split}
J_{AB}(f(\mathbf{x}))&=\E\left[l_{AB}(f(\mathbf{x}),y)\right]\\
&=\Prob\left(y=1|\mathbf{x}\right)\exp(-f(\mathbf{x}))+\Prob\left(y=-1|\mathbf{x}\right)\exp(f(\mathbf{x}))
\end{split}
\end{equation}

If we minimize that risk we will obtain the optimal predictor $f_{AB}(\mathbf{x})$, that turns out to be the symmetric logistic transform of $\Prob\left(y=1|\mathbf{x}\right)$.

\begin{equation}
\label{min_ab_eqn}
f_{AB}(\mathbf{x})=\dfrac{1}{2}\log\dfrac{\Prob\left(y=1|\mathbf{x}\right)}{\Prob\left(y=-1|\mathbf{x}\right)}
\end{equation}

\begin{figure}[!htb]
\centering
\includegraphics[width=7.5cm]{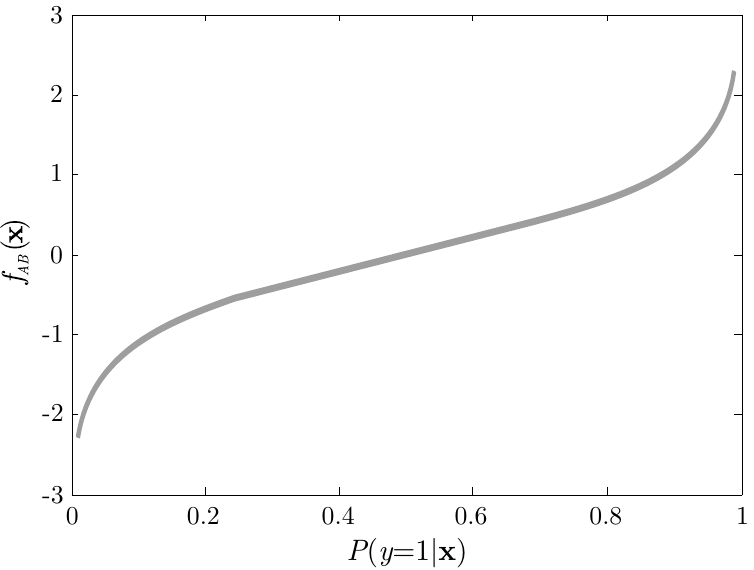}
\caption{Risk minimizing function (optimal predictor) for AdaBoost ($f_{AB}(\mathbf{x})$). It only depends on the likelihood of each class.}
\label{stat_cmp_ab_fig} 
\end{figure}

AdaBoost is geared to approximate, in an additive way, that optimal predictor without embedded costs. Thus, the obtained model will be cost-insensitive, only depending on the likelihood of each class (see Figure \ref{stat_cmp_ab_fig}).

In the case of Cost-Generalized AdaBoost, from this same perspective, we will have a loss function in which costs are included as modulators of the exponentials. 

\begin{equation}
\label{loss_cga_eqn}
l_{CGA}(f(\mathbf{x}),y)=\llbracket y=1 \rrbracket C_{P} \exp(-f(\mathbf{x})) + \llbracket y=-1 \rrbracket C_{N} \exp(f(\mathbf{x}))
\end{equation}

Thus, as explained in Section \ref{subsubsec:weight_cost_sensitive}, the respective risk function $J_{CGA}(f(\mathbf{x}))$ and its minimizer $f_{CGA}(\mathbf{x})$ will be the following ones:

\begin{equation}
\label{risk_cga_eqn}
\begin{split}
J_{CGA}(f(\mathbf{x}))&=\E\left[l_{CGA}(f(\mathbf{x}),y)\right]\\
&=\Prob\left(y=1|\mathbf{x}\right)C_{P}\exp(-f(\mathbf{x}))+\Prob\left(y=-1|\mathbf{x}\right)C_{N}\exp(f(\mathbf{x}))
\end{split}
\end{equation}

\begin{equation}
\label{min_cga_eqn}
f_{CGA}(\mathbf{x})=\dfrac{1}{2}\log\left(\dfrac{C_{P}\Prob\left(y=1|\mathbf{x}\right)}{C_{N}\Prob\left(y=-1|\mathbf{x}\right)}\right)
\end{equation}

As can be seen, now we have a cost-sensitive risk function with a cost-sensitive minimizer gearing to an optimal predictor $f_{CGA}(\mathbf{x})$ based on the asymmetric logistic transform of $\Prob\left(y=1|\mathbf{x}\right)$. Thus, in contrast to AdaBoost, the model pursued by Cost-Generalized AdaBoost will not exclusively depend on the likelihood of each class, but also on the related costs.

\begin{figure}[!htb]
\centering
\includegraphics[width=7.5cm]{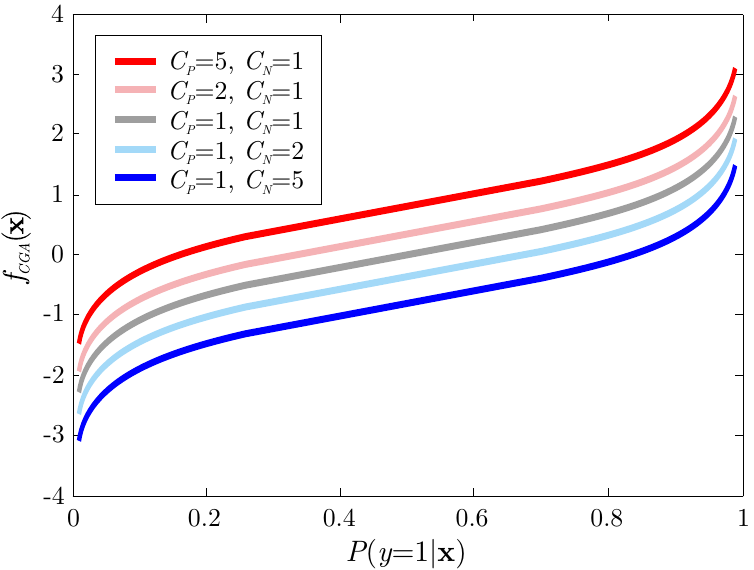}
\caption{Risk minimizing function (optimal predictor) for Cost-Generalized AdaBoost ($f_{CGA}(\mathbf{x})$). It depends on the likelihood of each class and on the related costs, having a homogeneous and continuous cost-sensitive behavior for whatever likelihood.}
\label{stat_cmp_cga_fig} 
\end{figure}

On the other hand, the loss function of Cost-Sensitive AdaBoost embeds the costs inside the exponents
\begin{equation}
\label{loss_csa_eqn}
l_{CSA}(f(\mathbf{x}),y)=\llbracket y=1 \rrbracket \exp(-C_{P} f(\mathbf{x})) + \llbracket y=-1 \rrbracket \exp(C_{N}f(\mathbf{x}))
\end{equation}
so the risk function and its associated minimizer will be as follows (see Section \ref{subsubsec:weight_cost_sensitive}):
\begin{equation}
\label{risk_csa_eqn}
\begin{split}
J_{CSA}(f(\mathbf{x}))&=\E\left[l_{CSA}(f(\mathbf{x}),y)\right]\\
&=\Prob\left(y=1|\mathbf{x}\right)\exp(-C_{P}f(\mathbf{x}))+\Prob\left(y=-1|\mathbf{x}\right)\exp(C_{N}f(\mathbf{x}))
\end{split}
\end{equation}
\begin{equation}
\label{min_csa_eqn}
f_{CSA}(\mathbf{x})=\dfrac{1}{C_{P}+C_{N}}\log\left(\dfrac{C_{P}\Prob\left(y=1|\mathbf{x}\right)}{C_{N}\Prob\left(y=-1|\mathbf{x}\right)}\right)
\end{equation}

\begin{figure}[!htb]
\centering
\includegraphics[width=7.5cm]{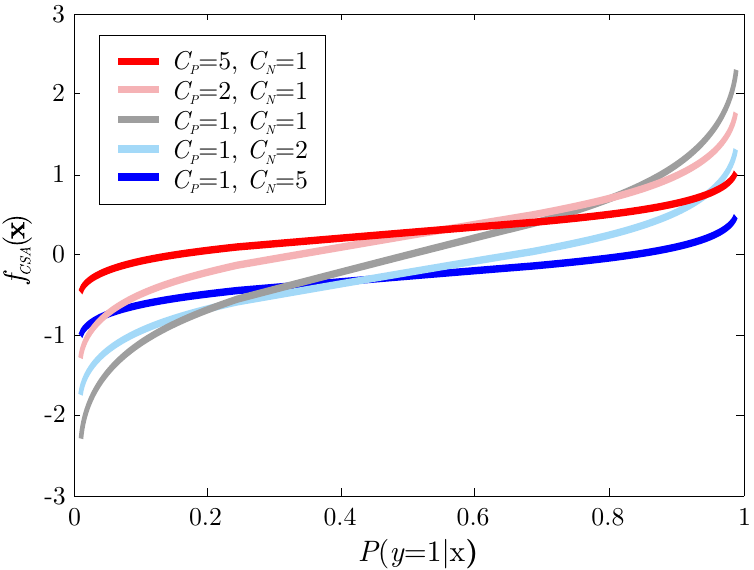}
\caption{Risk minimizing function (optimal predictor) for Cost-Sensitive AdaBoost ($f_{CSA}(\mathbf{x})$). It depends on the likelihood of each class and on the related costs, but in this case the cost-sensitive behavior is not homogeneous with respect to likelihood (solutions for different costs cross each other depending on $\Prob\left(y=1|\mathbf{x}\right)$).}
\label{stat_cmp_csa_fig} 
\end{figure}

Then, Cost-Sensitive AdaBoost is also aimed to fit a model based on the asymmetric logistic transform of $\Prob\left(y=1|\mathbf{x}\right)$, depending both on the likelihood of each class as well as on the related costs (see Figure \ref{stat_cmp_csa_fig}).

Notwithstanding, the optimal predictors guiding Cost-Sensitive AdaBoost and Cost-Generalized AdaBoost, despite being both cost-sensitive, have different equations. Such differences become apparent in their graphic representations (see Figures \ref{stat_cmp_cga_fig} and \ref{stat_cmp_csa_fig}).

\begin{figure}[!htb]
\centering
\subfloat[] 
{
    \label{training_set_nonover_fig}
    \includegraphics[width=5.5cm]{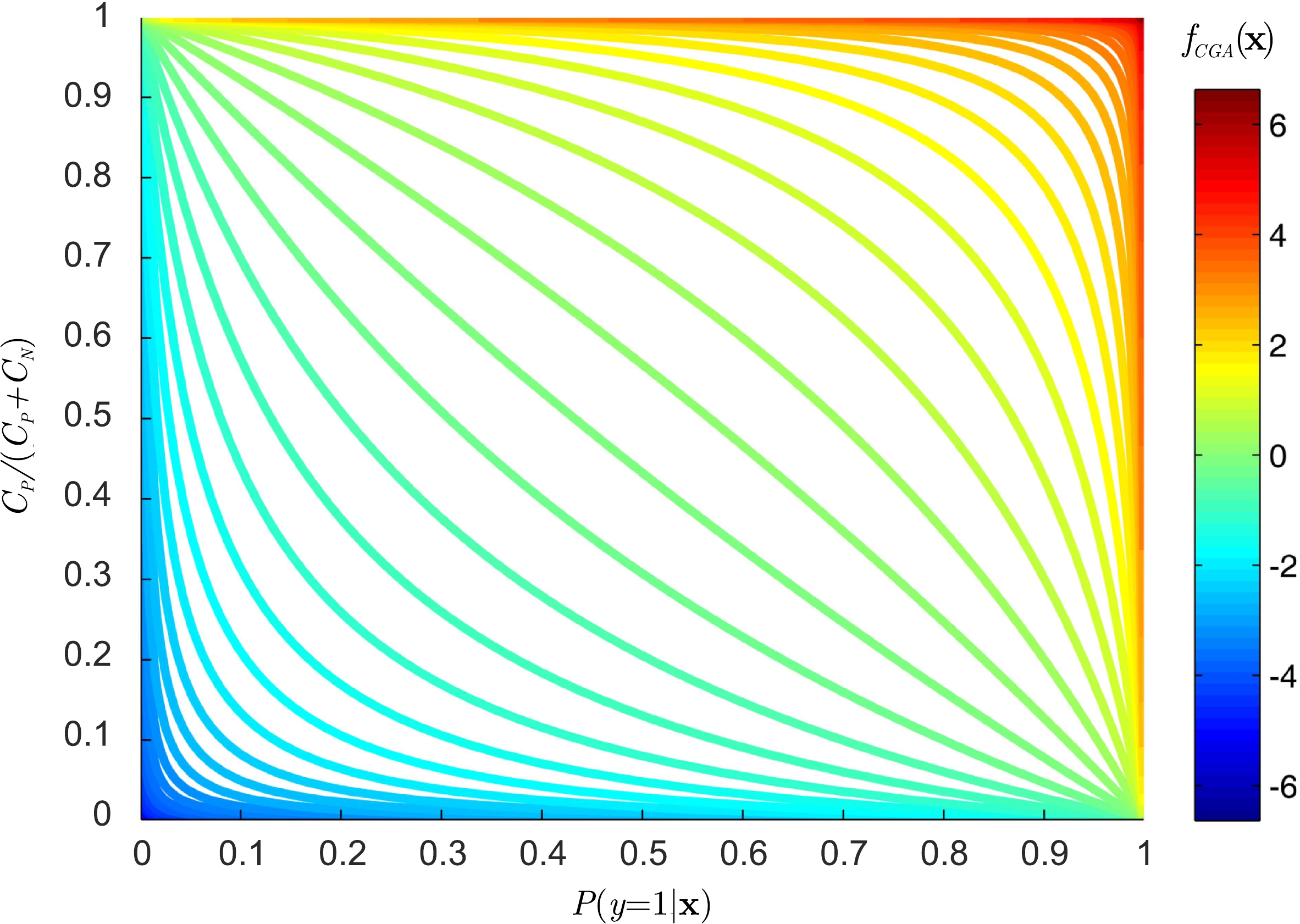}
}
\hspace{0.5cm}
\subfloat[] 
{
    \label{training_classifiers_nonover_fig}
    \includegraphics[width=5.5cm]{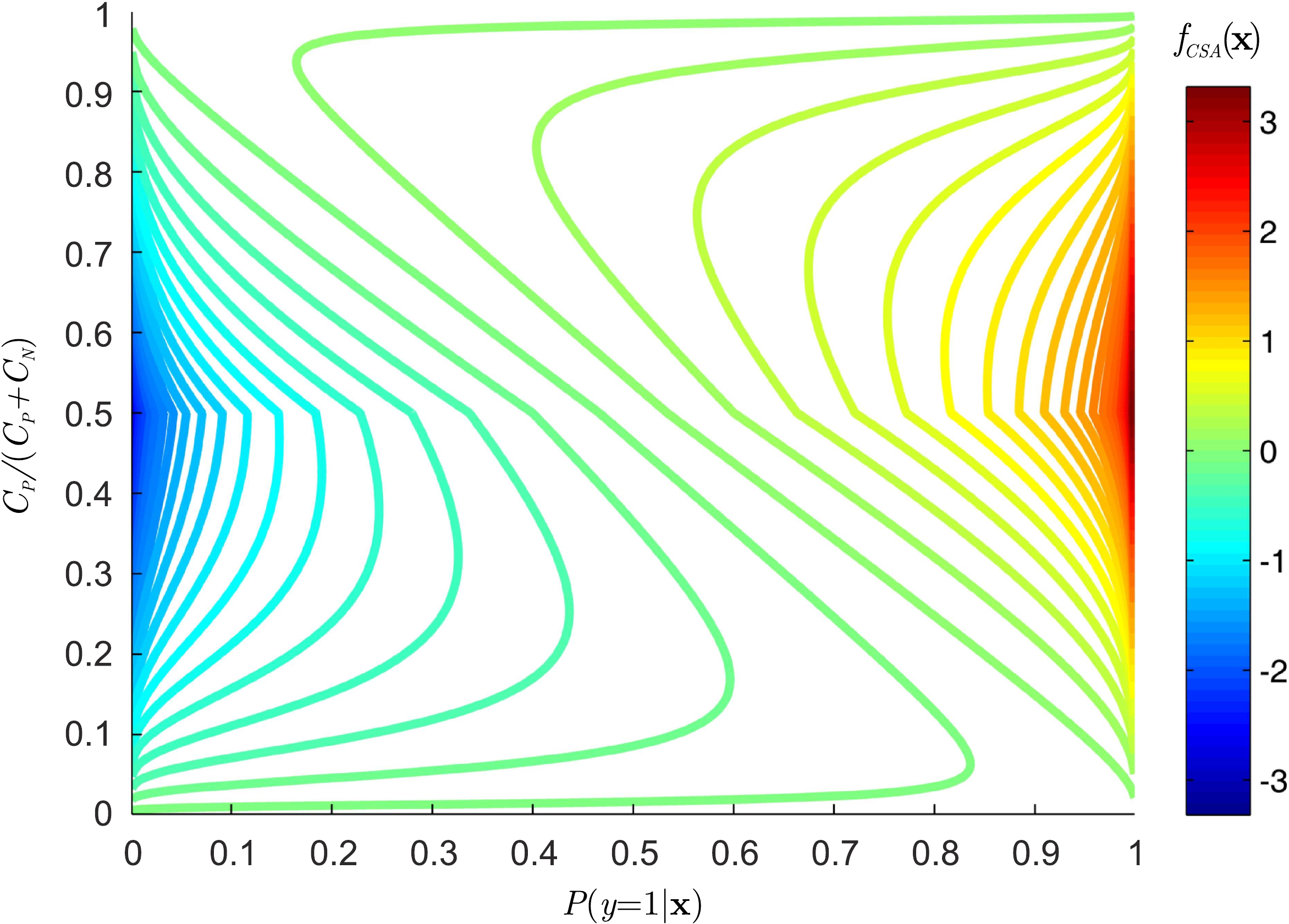}
}
\caption{Isolines of the optimal predictors for Cost-Generalized AdaBoost (a), and Cost-Sensitive AdaBoost (b), with respect to the likelihood ($\Prob\left(y=1|\mathbf{x}\right)$) and the normalized cost asymmetry ($\gamma=C_P/(C_P+C_N)$).}
\label{stat_cmp_cgacsamap_fig} 
\end{figure}

To delve into the consequences of these differences, we will analyze the optimal predictors of Cost-Generalized AdaBoost and Cost-Sensitive AdaBoost as functions depending on two magnitudes: likelihood and cost asymmetry \footnote{In the case of Cost-Sensitive AdaBoost (and AdaBoostDB) we can actually distinguish three different involved magnitudes (likelihood, cost of positives and cost of negatives), since the optimal predictor changes when costs are multiplied by a positive factor. This behavior (that does not happen for Cost-Generalized AdaBoost) violates the rules of the cost matrix \citep{Elkan01} explained at the beginning of Section \ref{sec:CSvar}. In order to tackle this problem for our analysis, we have restricted the possible costs to combinations $(C_P, C_N)$ in which one of the coefficients is always 1, and the other one is $\geq1$. This decision allows us to homogeneously interpret the scenarios in which negatives are the costliest class as label inversions.}. In Figure \ref{stat_cmp_cgacsamap_fig} we have represented the outputs of the optimal predictors as colormaps (we have used isolines for the sake of clarity) onto the plane defined by the likelihood and the cost asymmetry. As can be seen, the optimal predictor of Cost-Generalized AdaBoost (Figure \ref{stat_cmp_cgacsamap_fig}a) obtains higher predictor values for increasing $\Prob\left(y=1|\mathbf{x}\right)$ and increasing $C_P$ (vice versa for negatives). However, that is not the case for Cost-Sensitive AdaBoost (Figure \ref{stat_cmp_cgacsamap_fig}b) where, for a given likelihood, we can find lower predictor outputs for increasing positive costs (and vice versa for negatives). This inhomogeneous behavior can explain the \emph{asymmetry swapping} effect we have commented in Section \ref{subsubsec:error_bound_cmp}, and to which we will come back in the companion paper of the series \citep{LandesaAlba??b} when analyzing the experimental behavior of Cost-Sensitive AdaBoost.

\section{Summary and Conclusions}
\label{sec:Conclusions1}

In this first paper of the series we have introduced our working scenario, presenting the algorithms under study (AdaBoost with threshold modification \citep{ViolaJones04}; AsymBoost \citep{ViolaJones02}; AdaCost \citep{Fan99}; CSB0, CSB1 and CSB2 \citep{Ting98,Ting00}; AdaC1, AdaC2 and AdaC3 \citep{Sun05, Sun07}; Cost-Sensitive AdaBoost \citep{MasnadiVasconcelos07, MasnadiVasconcelos11}; AdaBoostDB \citep{LandesaAlba13}; and Cost-Generalized AdaBoost \citep{LandesaAlba12}) in a homogeneous notational framework and proposing a clustering scheme for them based on the way asymmetry is inserted in the learning process: \emph{theoretically}, \emph{heuristically} or \emph{a posteriori}. Then, for those algorithms with a fully theoretical derivation, we performed a thorough theoretical analysis and discussion, adopting the different perspectives that have been used to explain and derive the related approaches in the literature (Error Bound Minimization perspective \citep{SchapireSinger99} and Statistical View of Boosting \citep{Friedman00}).

The presented analysis clearly shows that the asymmetric weight initialization mechanism used by Cost-Generalized AdaBoost, from whatever point of view, is definitely a valid mechanism to build theoretically sound cost-sensitive boosted classifiers, despite having being recurrently overlooked or rejected in many previous works (e.g. \citep{Fan99, Ting00, ViolaJones02, MasnadiVasconcelos07, MasnadiVasconcelos11}). In addition, and besides being the simplest algorithm, Cost-Generalized AdaBoost exhibits the most consistent error bound definition and it is able to preserve the class-dependent loss ratio regardless of the training round whereas Cost-Sensitive AdaBoost and AdaBoostDB, the other theoretical alternatives, may end up emphasizing the least costly class.

After this purely theoretical study, an empirical analysis of the different approaches, also including the non-fully-theoretical methods (a posteriori and heuristic), is needed to reach global conclusions and culminate the analysis we have started in this paper. Such experimental part can be found in the next article of the series: ``\titlePartII'' \citep{LandesaAlba??b}.

%% file: Revisiting_Appendices1.tex


\newpage

\begin{appendices}

\phantomsection
\addcontentsline{toc}{section}{\protect\numberline{}Appendices}

%

\section{Unifying Compendium of Algorithms} 
\label{app:algorithms}

Next, a compendium of all the algorithms studied in this series of papers is presented, following an homogeneous and unifying notation to facilitate their comparative analysis.  Asterisks indicate changes with regard to standard AdaBoost.
\\
\\

\begin{algorithm}[H]

   \caption{AdaBoost}
   \label{adb_algorithm}
   \scriptsize
   \begin{algorithmic}
   \STATE {\bfseries Input:} \\Training set of $n$ examples: $(\mathbf{x_i}, y_i)$, where $y_{i}=$
   $\left\{
   \begin{array}{ll}
   1 & \mbox{if $1 \leq i \leq m$} \text{  (Positives)},\\
   -1 & \mbox{if $m < i \leq n$} \text{  (Negatives)}.
   \end{array} \right.$\\    
   Pool of $F$ weak classifiers: $h_{f}(\mathbf{x})$\\ 
   Number of rounds: $T$\\
   Initial (uniform) weight distribution: $D(i)$\\ 
   \vspace{4pt}
   \STATE {\bfseries Iterate:} \\
   \FOR{$t=1$ {\bfseries to} $T$}
   \FOR{$f=1$ {\bfseries to} $F$}
   \vspace{4pt}
   \STATE Compute the weighted error of the $f^{th}$ weak classifier, $\epsilon_{t,f}= \sum_{i=1}^{n} D(i) \llbracket h_{f}(\mathbf{x}_{i}) \neq y_{i}\rrbracket$
   \vspace{4pt}
   \ENDFOR
   \vspace{4pt}
   \STATE Select the weak learner $h_t(\mathbf{x})$ of smallest error, $\epsilon_{t}=\underset{f}{\operatorname{arg min}} \left[\epsilon_{t,f}\right]$
   \STATE Compute the goodness parameter of the selected classifier, $\alpha_t=\frac{1}{2}\log\left(\frac{1-\epsilon_{t}}{\epsilon_{t}}\right)$
   \STATE Update weights: $D(i)\leftarrow \frac{D(i)\exp\left(-\alpha_{t}y_{i}h_{t}\left(\mathbf{x}_i\right)\right)}{\sum_{i=1}^{n}D(i)\exp\left(-\alpha_{t}y_{i}h_{t}\left(\mathbf{x}_i\right)\right)}$
	 \vspace{4pt}
   \ENDFOR
   \vspace{4pt}
   \STATE {\bfseries Final Classifier:}\\
   \vspace{4pt}
    $H(\mathbf{x})=\mathrm{sign}\left(\sum_{t=1}^{T}\alpha_{t}h_{t}(\mathbf{x})\right)$
\end{algorithmic}
\end{algorithm}

\begin{algorithm}[p]
   \caption{AsymBoost}
   \label{asb_algorithm}
   \scriptsize
   \begin{algorithmic}
   \STATE {\bfseries Input:} \\Training set of $n$ examples: $(\mathbf{x_i}, y_i)$, where $y_{i}=$
   $\left\{
   \begin{array}{ll}
   1 & \mbox{if $1 \leq i \leq m$} \text{  (Positives)},\\
   -1 & \mbox{if $m < i \leq n$} \text{  (Negatives)}.
   \end{array} \right.$\\    
   Pool of $F$ weak classifiers: $h_{f}(\mathbf{x})$\\
   Number of rounds: $T$\\
   Initial (uniform) weight distribution: $D(i)$\\
   $\star$ Cost parameters: $C_{P},C_{N}\in\mathbb{R}^{+}$\\
   \vspace{4pt}
   \STATE {\bfseries Initialize:} \\ $\star$ Pre-emphasize weight distribution: $D(i)=\frac{D(i)\left(\frac{C_P}{C_N}\right)^{\frac{y_{i}}{2T}}}{\sum_{i=1}^{n}D(i)\left(\frac{C_P}{C_N}\right)^{\frac{y_{i}}{2T}}}$ 
   \vspace{4pt}
   \STATE {\bfseries Iterate:} \\
   \FOR{$t=1$ {\bfseries to} $T$}
   \FOR{$f=1$ {\bfseries to} $F$}
   \vspace{4pt}
   \STATE Compute the weighted error of the $f^{th}$ weak classifier, $\epsilon_{t,f}= \sum_{i=1}^{n} D(i) \llbracket h_{f}(\mathbf{x}_{i}) \neq y_{i}\rrbracket$
   \vspace{4pt}
   \ENDFOR
   \vspace{4pt}
   \STATE Select the weak learner $h_t(\mathbf{x})$ of smallest error, $\epsilon_{t}=\underset{f}{\operatorname{arg min}} \left[\epsilon_{t,f}\right]$
   \STATE Compute the goodness parameter of the selected classifier, $\alpha_t=\frac{1}{2}\log\left(\frac{1-\epsilon_{t}}{\epsilon_{t}}\right)$
   \STATE $\star$ Update weights: $D(i)\leftarrow \frac{D(i)\exp\left(-\alpha_{t}y_{i}h_{t}\left(\mathbf{x}_i\right)\right)\left(\frac{C_P}{C_N}\right)^{\frac{y_{i}}{2T}}}{\sum_{i=1}^{n}D(i)\exp\left(-\alpha_{t}y_{i}h_{t}\left(\mathbf{x}_i\right)\right)\left(\frac{C_P}{C_N}\right)^{\frac{y_{i}}{2T}}}$
	 \vspace{4pt}
   \ENDFOR
   \vspace{4pt}
   \STATE {\bfseries Final Classifier:}\\
   \vspace{4pt}
    $H(\mathbf{x})=\mathrm{sign}\left(\sum_{t=1}^{T}\alpha_{t}h_{t}(\mathbf{x})\right)$
\end{algorithmic}
\end{algorithm}

\clearpage

\begin{algorithm}[p]
   \caption{AdaCost}
   \label{adc_algorithm}
   \scriptsize
   \begin{algorithmic}
   \STATE {\bfseries Input:} \\Training set of $n$ examples: $(\mathbf{x_i}, y_i)$, where $y_{i}=$
   $\left\{
   \begin{array}{ll}
   1 & \mbox{if $1 \leq i \leq m$} \text{  (Positives)},\\
   -1 & \mbox{if $m < i \leq n$} \text{  (Negatives)}.
   \end{array} \right.$\\    
   Pool of $F$ weak classifiers: $h_{f}(\mathbf{x})$\\
   Number of rounds: $T$\\
	 Initial (uniform) weight distribution: $D(i)$\\
	 $\star$ Cost function: $C(i)\in(0,1)$ such as 
	 $\left\{
   \begin{array}{ll}
   \frac{C_{P}}{C_{P}+C_{N}} & \mbox{if $1 \leq i \leq m$},\\
   \frac{C_{N}}{C_{P}+C_{N}} & \mbox{if $m < i \leq n$}.
   \end{array} \right.$ with $C_{P},C_{N}\in\mathbb{R}^{+}$\\
   \vspace{4pt}
   \STATE {\bfseries Initialize:} \\ $\star$ Pre-emphasize weight distribution: $D(i)=\frac{C(i)D(i)}{\sum_{i=1}^{n}C(i)D(i)}$\\
   \vspace{4pt}
   \STATE {\bfseries Iterate:} \\
   \FOR{$t=1$ {\bfseries to} $T$}
   \FOR{$f=1$ {\bfseries to} $F$}
   \vspace{4pt}
   \STATE Compute the weighted error of the $f^{th}$ weak classifier, $\epsilon_{t,f}= \sum_{i=1}^{n} D(i) \llbracket h_{f}(\mathbf{x}_{i}) \neq y_{i}\rrbracket$
   \vspace{4pt}
   \ENDFOR
   \vspace{4pt}
   \STATE Select the weak learner $h_t(\mathbf{x})$ of smallest error, $\epsilon_{t}=\underset{f}{\operatorname{arg min}} \left[\epsilon_{t,f}\right]$
   \STATE $\star$ Compute the cost-adjustment function, $\beta(i)=$
   $\left\{
   \begin{array}{ll}
   0.5\left(1-C(i)\right) & \mbox{$\text{if } h_{f}(\mathbf{x}_{i}) = y_{i}$},\\
   0.5\left(1+C(i)\right) & \mbox{$\text{if } h_{f}(\mathbf{x}_{i}) \neq y_{i}$}.
   \end{array} \right.$\\
   \STATE $\star$ Compute the goodness of the selected classifier with cost-adjustment as \\
   $\alpha_t=\frac{1}{2}\log\left(\frac{1+\sum_{i=1}^{n}D(i)y_{i}h_{t}\beta(i)}{1-\sum_{i=1}^{n}D(i)y_{i}h_{t}\beta(i)}\right)$
   \STATE $\star$ Update weights: $D(i)\leftarrow \frac{D(i)\exp\left(-\alpha_{t}y_{i}h_{t}\left(\mathbf{x}_i\right)\beta(i)\right)}{\sum_{i=1}^{n}D(i)\exp\left(-\alpha_{t}y_{i}h_{t}\left(\mathbf{x}_i\right)\beta(i)\right)}$
	 \vspace{4pt}
   \ENDFOR
   \vspace{4pt}
   \STATE {\bfseries Final Classifier:}\\
   \vspace{4pt}
    $H(\mathbf{x})=\mathrm{sign}\left(\sum_{t=1}^{T}\alpha_{t}h_{t}(\mathbf{x})\right)$
\end{algorithmic}
\end{algorithm}

\begin{algorithm}[p]
   \caption{CSB0}
   \label{csb0_algorithm}
   \scriptsize
   \begin{algorithmic}
   \STATE {\bfseries Input:} \\Training set of $n$ examples: $(\mathbf{x_i}, y_i)$, where 
   \begin{math}
   y_{i}= \left\{
   \begin{array}{ll}
   1 & \mbox{if $1 \leq i \leq m$} \text{  (Positives)},\\
   -1 & \mbox{if $m < i \leq n$} \text{  (Negatives)}.
   \end{array} \right.
   \end{math}\\
   Pool of $F$ weak classifiers: $h_{f}(\mathbf{x})$\\ 
   Number of rounds: $T$\\
   $\star$ Cost parameters: $C_{P},C_{N}\in\mathbb{R}^{+}$\\
	\STATE {\bfseries Initialize:} \\
   $\star$ Weight Distribution: \begin{math}
   D(i)= \left\{
   \begin{array}{ll}
   \frac{C_{P}}{mC_{P}+(n-m)C_{N}} & \mbox{if $1 \leq i \leq m$},\\
   \frac{C_{N}}{mC_{P}+(n-m)C_{N}} & \mbox{if $m < i \leq n$}.
   \end{array} \right.
   \end{math} \\
   \vspace{4pt}
   \STATE {\bfseries Iterate:} \\
   \FOR{$t=1$ {\bfseries to} $T$}
   \FOR{$f=1$ {\bfseries to} $F$}
   \vspace{4pt}
    \STATE Compute the weighted error of the $f^{th}$ weak classifier, $\epsilon_{t,f}= \sum_{i=1}^{n} D(i) \llbracket h_{f}(\mathbf{x}_{i}) \neq y_{i}\rrbracket$
   \vspace{4pt}
   \ENDFOR
   \vspace{4pt}
   \STATE Select the weak learner $h_t(\mathbf{x})$ of smallest error, $\epsilon_{t}=\underset{f}{\operatorname{arg min}} \left[\epsilon_{t,f}\right]$
   \STATE Compute the goodness parameter of the selected classifier, $\alpha_t=\frac{1}{2}\log\left(\frac{1-\epsilon_{t}}{\epsilon_{t}}\right)$
   \STATE $\star$ Update and normalize weights:\\
   \hspace{\algorithmicindent}\hspace{\algorithmicindent}\begin{math} 
   D(i)\leftarrow \left\{
   \begin{array}{ll}
   D(i) & \mbox{if $h_f(\mathbf{x}_{i})=y_i$},\\
   C_PD(i) & \mbox{if $h_f(\mathbf{x}_{i})\neq y_i$ and $y_i=1$},\\
   C_ND(i) & \mbox{if $h_f(\mathbf{x}_{i})\neq y_i$ and $y_i=-1$}.
   \end{array}
	 \right. 
   \end{math}\\
	 \hspace{\algorithmicindent}\hspace{\algorithmicindent}\begin{math}
	 D(i) \leftarrow \frac{D(i)}{\sum_{i=1}^{n}D(i)}
   \end{math}\\
	 \vspace{4pt}
   \ENDFOR
   \vspace{4pt}
   \STATE {\bfseries Final Classifier:}\\
   \vspace{4pt}
   $\star H(\mathbf{x})=\mathrm{sign}\left(\sum_{t=1}^{T}\alpha_{t}h_{t}(\mathbf{x}) \left( C_P \llbracket h_{t}(\mathbf{x})=+1 \rrbracket + C_N \llbracket h_{t}(\mathbf{x})=-1 \rrbracket \right)\right)$
\end{algorithmic}
\end{algorithm}

\begin{algorithm}[p]
   \caption{CSB1}
   \label{csb1_algorithm}
   \scriptsize
   \begin{algorithmic}
   \STATE {\bfseries Input:} \\Training set of $n$ examples: $(\mathbf{x_i}, y_i)$, where 
   \begin{math}
   y_{i}= \left\{
   \begin{array}{ll}
   1 & \mbox{if $1 \leq i \leq m$} \text{  (Positives)},\\
   -1 & \mbox{if $m < i \leq n$} \text{  (Negatives)}.
   \end{array} \right.
   \end{math}\\
   Pool of $F$ weak classifiers: $h_{f}(\mathbf{x})$\\ 
   Number of rounds: $T$\\
   $\star$ Cost parameters: $C_{P},C_{N}\in\mathbb{R}^{+}$\\
	\STATE {\bfseries Initialize:} \\
   $\star$ Weight Distribution: \begin{math}
   D(i)= \left\{
   \begin{array}{ll}
   \frac{C_{P}}{mC_{P}+(n-m)C_{N}} & \mbox{if $1 \leq i \leq m$},\\
   \frac{C_{N}}{mC_{P}+(n-m)C_{N}} & \mbox{if $m < i \leq n$}.
   \end{array} \right.
   \end{math} \\
   \vspace{4pt}
   \STATE {\bfseries Iterate:} \\
   \FOR{$t=1$ {\bfseries to} $T$}
   \FOR{$f=1$ {\bfseries to} $F$}
   \vspace{4pt}
    \STATE Compute the weighted error of the $f^{th}$ weak classifier, $\epsilon_{t,f}= \sum_{i=1}^{n} D(i) \llbracket h_{f}(\mathbf{x}_{i}) \neq y_{i}\rrbracket$
   \vspace{4pt}
   \ENDFOR
   \vspace{4pt}
   \STATE Select the weak learner $h_t(\mathbf{x})$ of smallest error, $\epsilon_{t}=\underset{f}{\operatorname{arg min}} \left[\epsilon_{t,f}\right]$
   \STATE Compute the goodness parameter of the selected classifier, $\alpha_t=\frac{1}{2}\log\left(\frac{1-\epsilon_{t}}{\epsilon_{t}}\right)$
   \STATE $\star$ Update and normalize weights:\\
   \hspace{\algorithmicindent}\hspace{\algorithmicindent}\begin{math} 
   D(i) \leftarrow \left\{
   \begin{array}{ll}
   D(i)\exp\left(-y_ih_f(\mathbf{x}_{i})\right) & \mbox{if $h_f(\mathbf{x}_{i})=y_i$},\\
   C_PD(i)\exp\left(-y_ih_f(\mathbf{x}_{i})\right) & \mbox{if $h_f(\mathbf{x}_{i})\neq y_i$ and $y_i=1$},\\
   C_ND(i)\exp\left(-y_ih_f(\mathbf{x}_{i})\right) & \mbox{if $h_f(\mathbf{x}_{i})\neq y_i$ and $y_i=-1$}.
   \end{array}
	 \right. 
   \end{math}\\
	  \hspace{\algorithmicindent}\hspace{\algorithmicindent}\begin{math} 
		 D(i) \leftarrow \frac{D(i)}{\sum_{i=1}^{n}D(i)}
		\end{math}\\
	 \vspace{4pt}
   \ENDFOR
   \vspace{4pt}
   \STATE {\bfseries Final Classifier:}\\
   \vspace{4pt}
   $\star H(\mathbf{x})=\mathrm{sign}\left(\sum_{t=1}^{T}\alpha_{t}h_{t}(\mathbf{x}) \left( C_P \llbracket h_{t}(\mathbf{x})=+1 \rrbracket + C_N \llbracket h_{t}(\mathbf{x})=-1 \rrbracket \right)\right)$
\end{algorithmic}
\end{algorithm}

\begin{algorithm}[p]
   \caption{CSB2}
   \label{csb2_algorithm}
   \scriptsize
   \begin{algorithmic}
   \STATE {\bfseries Input:} \\Training set of $n$ examples: $(\mathbf{x_i}, y_i)$, where 
   \begin{math}
   y_{i}= \left\{
   \begin{array}{ll}
   1 & \mbox{if $1 \leq i \leq m$} \text{  (Positives)},\\
   -1 & \mbox{if $m < i \leq n$} \text{  (Negatives)}.
   \end{array} \right.
   \end{math}\\
   Pool of $F$ weak classifiers: $h_{f}(\mathbf{x})$\\ 
   Number of rounds: $T$\\
   $\star$ Cost parameters: $C_{P},C_{N}\in\mathbb{R}^{+}$\\
	\STATE {\bfseries Initialize:} \\
   $\star$ Weight Distribution: \begin{math}
   D(i)= \left\{
   \begin{array}{ll}
   \frac{C_{P}}{mC_{P}+(n-m)C_{N}} & \mbox{if $1 \leq i \leq m$},\\
   \frac{C_{N}}{mC_{P}+(n-m)C_{N}} & \mbox{if $m < i \leq n$}.
   \end{array} \right.
   \end{math} \\
   \vspace{4pt}
   \STATE {\bfseries Iterate:} \\
   \FOR{$t=1$ {\bfseries to} $T$}
   \FOR{$f=1$ {\bfseries to} $F$}
   \vspace{4pt}
    \STATE Compute the weighted error of the $f^{th}$ weak classifier, $\epsilon_{t,f}= \sum_{i=1}^{n} D(i) \llbracket h_{f}(\mathbf{x}_{i}) \neq y_{i}\rrbracket$
   \vspace{4pt}
   \ENDFOR
   \vspace{4pt}
   \STATE Select the weak learner $h_t(\mathbf{x})$ of smallest error, $\epsilon_{t}=\underset{f}{\operatorname{arg min}} \left[\epsilon_{t,f}\right]$
   \STATE Compute the goodness parameter of the selected classifier, $\alpha_t=\frac{1}{2}\log\left(\frac{1-\epsilon_{t}}{\epsilon_{t}}\right)$
   \STATE $\star$ Update and normalize weights:\\
   \hspace{\algorithmicindent}\hspace{\algorithmicindent}\begin{math} 
   D(i) \leftarrow \left\{
   \begin{array}{ll}
   D(i)\exp\left(-\alpha_ty_ih_f(\mathbf{x}_{i})\right) & \mbox{if $h_f(\mathbf{x}_{i})=y_i$},\\
   C_PD(i)\exp\left(-\alpha_ty_ih_f(\mathbf{x}_{i})\right) & \mbox{if $h_f(\mathbf{x}_{i})\neq y_i$ and $y_i=1$},\\
   C_ND(i)\exp\left(-\alpha_ty_ih_f(\mathbf{x}_{i})\right) & \mbox{if $h_f(\mathbf{x}_{i})\neq y_i$ and $y_i=-1$}.
   \end{array} 
	 \right. 
   \end{math}\\
	\hspace{\algorithmicindent}\hspace{\algorithmicindent}\begin{math} 
	D(i) \leftarrow \frac{D(i)}{\sum_{i=1}^{n}D(i)}
	\end{math}\\
	 \vspace{4pt}
   \ENDFOR
   \vspace{4pt}
   \STATE {\bfseries Final Classifier:}\\
   \vspace{4pt}
   $\star H(\mathbf{x})=\mathrm{sign}\left(\sum_{t=1}^{T}\alpha_{t}h_{t}(\mathbf{x}) \left( C_P \llbracket h_{t}(\mathbf{x})=+1 \rrbracket + C_N \llbracket h_{t}(\mathbf{x})=-1 \rrbracket \right)\right)$
\end{algorithmic}
\end{algorithm}

\begin{algorithm}[p]
   \caption{AdaC1}
   \label{ac1_algorithm}
   \scriptsize
   \begin{algorithmic}
   \STATE {\bfseries Input:} \\Training set of $n$ examples: $(\mathbf{x_i}, y_i)$, where $y_{i}=$
   $\left\{
   \begin{array}{ll}
   1 & \mbox{if $1 \leq i \leq m$} \text{  (Positives)},\\
   -1 & \mbox{if $m < i \leq n$} \text{  (Negatives)}.
   \end{array} \right.$\\    
   Pool of $F$ weak classifiers: $h_{f}(\mathbf{x})$\\ 
   Number of rounds: $T$\\
   Initial (uniform) weight distribution: $D(i)$\\ 
   $\star$ Cost function: $C(i)\in (0,1)$ such as 
	 $\left\{
   \begin{array}{ll}
   C_{P} & \mbox{if $1 \leq i \leq m$},\\
   C_{N} & \mbox{if $m < i \leq n$}.
   \end{array} \right.$\\
   \vspace{4pt}
   \STATE {\bfseries Iterate:} \\
   \FOR{$t=1$ {\bfseries to} $T$}
    \STATE $\star$ Compute the total weight with costs, $W_{t}= \sum_{i=1}^{n} c(i)D(i)$
   \FOR{$f=1$ {\bfseries to} $F$}
   \vspace{4pt}
   \STATE $\star$ Compute the weighted error of the $f^{th}$ weak classifier with costs,\\ $\epsilon_{t,f}= \sum_{i=1}^{n} c(i)D(i) \llbracket h_{f}(\mathbf{x}_{i}) \neq y_{i}\rrbracket$
   \vspace{4pt}
   \ENDFOR
   \vspace{4pt}
   \STATE Select the weak learner $h_t(\mathbf{x})$ of smallest error, $\epsilon_{t}=\underset{f}{\operatorname{arg min}} \left[\epsilon_{t,f}\right]$
   \STATE $\star$ Compute the goodness parameter of the selected classifier, $\alpha_t=\frac{1}{2}\log\left(\frac{1+W_{t}-2\epsilon_{t}}{1-W_{t}+2\epsilon_{t}}\right)$
   \STATE $\star$ Update weights: $D(i) \leftarrow \frac{D(i)\exp\left(-\alpha_{t}c_{i}y_{i}h_{t}\left(\mathbf{x}_i\right)\right)}{\sum_{i=1}^{n}D(i)\exp\left(-\alpha_{t}c_{i}y_{i}h_{t}\left(\mathbf{x}_i\right)\right)}$
	 \vspace{4pt}
   \ENDFOR
   \vspace{4pt}
   \STATE {\bfseries Final Classifier:}\\
   \vspace{4pt}
    $H(\mathbf{x})=\mathrm{sign}\left(\sum_{t=1}^{T}\alpha_{t}h_{t}(\mathbf{x})\right)$
\end{algorithmic}
\end{algorithm}

\begin{algorithm}[p]
   \caption{AdaC2}
   \label{ac2_algorithm}
   \scriptsize
   \begin{algorithmic}
   \STATE {\bfseries Input:} \\Training set of $n$ examples: $(\mathbf{x_i}, y_i)$, where $y_{i}=$
   $\left\{
   \begin{array}{ll}
   1 & \mbox{if $1 \leq i \leq m$} \text{  (Positives)},\\
   -1 & \mbox{if $m < i \leq n$} \text{  (Negatives)}.
   \end{array} \right.$\\    
   Pool of $F$ weak classifiers: $h_{f}(\mathbf{x})$\\ 
   Number of rounds: $T$\\
   Initial (uniform) weight distribution: $D(i)$\\ 
   $\star$ Cost function: $C(i)\in (0,1)$ such as 
	 $\left\{
   \begin{array}{ll}
   C_{P} & \mbox{if $1 \leq i \leq m$},\\
   C_{N} & \mbox{if $m < i \leq n$}.
   \end{array} \right.$\\
   \vspace{4pt}
   \STATE {\bfseries Iterate:} \\
   \FOR{$t=1$ {\bfseries to} $T$}
    \STATE $\star$ Compute the total weight with costs, $W_{t}= \sum_{i=1}^{n} c(i)D(i)$
   \FOR{$f=1$ {\bfseries to} $F$}
   \vspace{4pt}
   \STATE $\star$ Compute the weighted error of the $f^{th}$ weak classifier with costs,\\ $\epsilon_{t,f}= \sum_{i=1}^{n} c(i)D(i) \llbracket h_{f}(\mathbf{x}_{i}) \neq y_{i}\rrbracket$
   \vspace{4pt}
   \ENDFOR
   \vspace{4pt}
   \STATE Select the weak learner $h_t(\mathbf{x})$ of smallest error, $\epsilon_{t}=\underset{f}{\operatorname{arg min}} \left[\epsilon_{t,f}\right]$
   \STATE $\star$ Compute the goodness parameter of the selected classifier, $\alpha_t=\frac{1}{2}\log\left(\frac{W_{t}-\epsilon_{t}}{\epsilon_{t}}\right)$
   \STATE $\star$ Update weights: $D(i) \leftarrow \frac{c_{i}D(i)\exp\left(-\alpha_{t}y_{i}h_{t}\left(\mathbf{x}_i\right)\right)}{\sum_{i=1}^{n}c_{i}D(i)\exp\left(-\alpha_{t}y_{i}h_{t}\left(\mathbf{x}_i\right)\right)}$
	 \vspace{4pt}
   \ENDFOR
   \vspace{4pt}
   \STATE {\bfseries Final Classifier:}\\
   \vspace{4pt}
    $H(\mathbf{x})=\mathrm{sign}\left(\sum_{t=1}^{T}\alpha_{t}h_{t}(\mathbf{x})\right)$
\end{algorithmic}
\end{algorithm}

\begin{algorithm}[p]
   \caption{AdaC3}
   \label{ac3_algorithm}
   \scriptsize
   \begin{algorithmic}
   \STATE {\bfseries Input:} \\Training set of $n$ examples: $(\mathbf{x_i}, y_i)$, where $y_{i}=$
   $\left\{
   \begin{array}{ll}
   1 & \mbox{if $1 \leq i \leq m$} \text{  (Positives)},\\
   -1 & \mbox{if $m < i \leq n$} \text{  (Negatives)}.
   \end{array} \right.$\\    
   Pool of $F$ weak classifiers: $h_{f}(\mathbf{x})$\\ 
   Number of rounds: $T$\\
   Initial (uniform) weight distribution: $D(i)$\\ 
   $\star$ Cost function: $C(i)\in (0,1)$ such as 
	 $\left\{
   \begin{array}{ll}
   C_{P} & \mbox{if $1 \leq i \leq m$},\\
   C_{N} & \mbox{if $m < i \leq n$}.
   \end{array} \right.$\\
   \vspace{4pt}
   \STATE {\bfseries Iterate:} \\
   \FOR{$t=1$ {\bfseries to} $T$}
    \STATE $\star$ Compute the total weight with costs, $W^{1}_{t}= \sum_{i=1}^{n} c(i)D(i)$
    \STATE $\star$ Compute the total weight with squared costs, $W^{2}_{t}= \sum_{i=1}^{n} c(i)^{2}D(i)$
   \FOR{$f=1$ {\bfseries to} $F$}
   \vspace{4pt}
   \STATE $\star$ Compute the weighted error of the $f^{th}$ weak classifier with squared costs,\\ $\epsilon_{t,f}= \sum_{i=1}^{n} c(i)^2D(i) \llbracket h_{f}(\mathbf{x}_{i}) \neq y_{i}\rrbracket$
   \vspace{4pt}
   \ENDFOR
   \vspace{4pt}
   \STATE Select the weak learner $h_t(\mathbf{x})$ of smallest error, $\epsilon_{t}=\underset{f}{\operatorname{arg min}} \left[\epsilon_{t,f}\right]$
   \STATE $\star$ Compute the goodness parameter of the selected classifier, $\alpha_t=\frac{1}{2}\log\left(\frac{W^{1}_{t}+W^{2}_{t}-2\epsilon_{t}}{W^{1}_{t}-W^{2}_t+2\epsilon_{t}}\right)$
   \STATE $\star$ Update weights: $D(i) \leftarrow \frac{c_{i}D(i)\exp\left(-\alpha_{t}c_{i}y_{i}h_{t}\left(\mathbf{x}_i\right)\right)}{\sum_{i=1}^{n}c_{i}D(i)\exp\left(-\alpha_{t}c_{i}y_{i}h_{t}\left(\mathbf{x}_i\right)\right)}$
	 \vspace{4pt}
   \ENDFOR
   \vspace{4pt}
   \STATE {\bfseries Final Classifier:}\\
   \vspace{4pt}
    $H(\mathbf{x})=\mathrm{sign}\left(\sum_{t=1}^{T}\alpha_{t}h_{t}(\mathbf{x})\right)$
\end{algorithmic}
\end{algorithm}

\begin{algorithm}[p]
   \caption{Cost-Sensitive AdaBoost}
   \label{csa_algorithm}
   \scriptsize
   \begin{algorithmic}
   \STATE {\bfseries Input:} \\Training set of $n$ examples: $(\mathbf{x_i}, y_i)$, where $y_{i}=$ 
   $\left\{
   \begin{array}{ll}
   1 & \mbox{if $1 \leq i \leq m$},\\
   -1 & \mbox{if $m < i \leq n$}.
   \end{array} \right.$\\   
   Pool of $F$ weak classifiers: $h_{f}(\mathbf{x})$\\ $\star$ Cost parameters: $C_{P}$, $C_{N} \in \mathbb{R}^{+}$\\ Number of rounds: $T$\\
   Initial (uniform) weight distribution: $D(i)$
   \vspace{4pt}
   \FOR{$t=1$ {\bfseries to} $T$}
   \vspace{4pt}
   \STATE $\star$ Calculate parameters:\\
   $\mathcal{T}_{P}=\sum_{i=1}^{m}D(i)$\\
   $\mathcal{T}_{N}=\sum_{i=m+1}^{n}D(i)$\\
   \vspace{4pt}
   \FOR{$f=1$ {\bfseries to} $F$}
   \vspace{4pt}
   \STATE Pick up $f^{th}$ weak classifier: $h_{f}(\mathbf{x})$.\\
   \vspace{4pt}
   \STATE $\star$ Calculate parameters:\\
   $\begin{array}{ll}
   \mathcal{B}=\sum_{i=1}^{m}D(i)\llbracket y_{i} \neq h_{f}(\mathbf{x_{i}})\rrbracket,\\
   \mathcal{D}=\sum_{i=m+1}^{n}D(i)\llbracket y_{i} \neq h_{f}(\mathbf{x_{i}})\rrbracket.
   \end{array}.$\\
   
   \vspace{4pt}
   \STATE $\star$ Find $\alpha_{t,f}$ solving the next hyperbolic equation:\\
   $2C_{P}\mathcal{B}\cosh\left(C_{P}\alpha_{t,f}\right)+2C_{N}\mathcal{D}\cosh\left(C_{N}\alpha_{t,f}\right)=C_{1}\mathcal{T}_{P}\mathrm{e}^{-C_{P}\alpha{t,f}}+C_{2}\mathcal{T}_{N}\mathrm{e}^{-C_{N}\alpha_{t,f}}$\\
   \vspace{4pt}
   \STATE $\star$ Compute the loss of the weak learner\\ $L_{t,f}=\mathcal{B}\left(\mathrm{e}^{C_{P}\alpha_{t,f}}-\mathrm{e}^{-C_{P}\alpha_{t,f}}\right)+\mathcal{T}_{P}\mathrm{e}^{-C_{P}\alpha_{t,f}} +\mathcal{D}\left(\mathrm{e}^{C_{N}\alpha_{t,f}}-\mathrm{e}^{-C_{N}\alpha_{t,f}}\right) +\mathcal{T}_{N}\mathrm{e}^{-C_{N}\alpha_{t,f}}$
 	\vspace{4pt}
   \ENDFOR
   \vspace{4pt}
   \STATE Select the weak learner $\left(h_t(\mathbf{x}), \alpha_{t}(\mathbf{x})\right)$ of smallest loss in this round:  $\underset{f}{\operatorname{arg min}} \left[L_{t,f}\right]$\\
   \STATE $\star$ Update weights: \\ 
   $D(i) \leftarrow \left\{
   \begin{array}{ll}
   D(i)\exp\left(-C_{P}\alpha_{t}h_{t}\left(\mathbf{x_i}\right)\right) & \mbox{if $1 \leq i \leq m$},\\
   D(i)\exp\left(C_{N}\alpha_{t}h_{t}\left(\mathbf{x_i}\right)\right) & \mbox{if $m < i \leq n$}.
   \end{array} \right.$\\
   \vspace{4pt}
   \ENDFOR
   \vspace{4pt}
   \STATE {\bfseries Final Classifier:}\\
   \vspace{4pt}
    $H(\mathbf{x})=\mathrm{sign}\left(\sum_{t=1}^{T}\alpha_{t}h_{t}(\mathbf{x})\right)$
\end{algorithmic}
\end{algorithm}

\begin{algorithm}[p]
   \caption{AdaBoostDB}
   \label{abdb_algorithm}
   \scriptsize
\begin{algorithmic}
   \STATE {\bfseries Input:} \\Training set of $n$ examples: $(\mathbf{x_i}, y_i)$, where $y_{i}= \left\{
   \begin{array}{cl}
   1 & \mbox{if $1 \leq i \leq m$},\\
   -1 & \mbox{if $m < i \leq n$}.
   \end{array} \right. $\\ Pool of $F$ weak classifiers: $h_{f}(\mathbf{x})$\\ $\star$ Cost parameters: $C_{P}$, $C_{N} \in \mathbb{R}^{+}$\\ Number of rounds: $T$\\
   Initial weight distribution: $D(i)$
\vspace{4pt}
   \STATE {\bfseries Initialize:} \\ $\star$ Weight subdistributions:
   $\left\{
   \begin{array}{ll}
   D_{P}(i)=\frac{D(i)}{\sum_{i=1}^{m}D(i)} & \mbox{if $1 \leq i \leq m$},\\  			   
   D_{N}(i)=\frac{D(i)}{\sum_{i=m+1}^{n}D(i)} & \mbox{if $m < i \leq n$}.
   \end{array} \right.$\\
   $\star$ Accumulators: $A_{P}=1$, $A_{N}=1$.\\
   \vspace{4pt}
   \FOR{$t=1$ {\bfseries to} $T$}
   \vspace{4pt}
   \STATE {\bfseries Initialize:} \\$\star$ Minimum root: $r=1$\\ $\star$ Minimum root vector: $\vec{r}=(2,2)$\\ $\star$ Scalar product: $s=1$
      \vspace{4pt}
   \STATE $\star$ Update accumulators:
   $\left\{
   \begin{array}{ll}
   A_{P} \leftarrow A_{P}\sum_{i}D_{P}(i),\\
   A_{N} \leftarrow A_{N}\sum_{i}D_{N}(i).
   \end{array} \right.$\\

   \STATE $\star$ Normalize weight subdistributions:
   $\left\{
   \begin{array}{ll}
    D_{P}(i) \leftarrow \frac{D_{P}(i)}{\sum_{i=1}^{m}D_{P}(i)},\\
   D_{N}(i) \leftarrow \frac{D_{N}(i)}{\sum_{i=m+1}^{n}D_{N}(i)}.
   \end{array} \right.$\\
   
   \STATE $\star$ Calculate static parameters: 
   $\left\{
   \begin{array}{ll}
   a=\frac{C_{P}A_{P}}{C_{P}A_{P}+C_{N}A_{N}},\\
   b=\frac{C_{N}A_{N}}{C_{P}A_{P}+C_{N}A_{N}}.
   \end{array} \right.$\\
   
      \vspace{4pt}
   \FOR{$f=1$ {\bfseries to} $F$}
      \vspace{4pt}
   \STATE $\star$ Calculate variable parameters:
   $\left\{
   \begin{array}{ll}
   \varepsilon_{P,f}=\sum_{i=1}^{m}D_{P}(i)\llbracket y_{i} \neq h_{f}(\mathbf{x_{i}})\rrbracket,\\
   \varepsilon_{N,f}=\sum_{i=m+1}^{n}D_{N}(i)\llbracket y_{i} \neq h_{f}(\mathbf{x_{i}})\rrbracket.
   \end{array} \right.$\\
   \vspace{4pt}
   \STATE $\star$ Calculate current classifier vector: $\vec{c}=(a\cdot\varepsilon_{P,f},b\cdot\varepsilon_{N,f})$\\
   \vspace{4pt}
   
 \fbox
 {\parbox{10cm} {
   \textbf{$\star$ CONDITIONAL SEARCH}
   \vspace{2pt}
	 \IF {$a\cdot\varepsilon_{P,f}+b\cdot\varepsilon_{N,f}<\frac{1}{2}$ [\textbf{Contribution Condition}]}
   \vspace{4pt}
   \IF {$\vec{c}\cdot\vec{r}>s$ [\textbf{Improvement Condition}]}
   \vspace{4pt}
   \STATE Search the only real and positive root $r$ of  the polynomial:\\
    $(a\cdot\varepsilon_{P,f})x^{2C_{P}}+ (b\cdot\varepsilon_{N,f})x^{C_{P}+C_{N}}+b(\varepsilon_{N,f}-1)x^{C_{P}-C_{N}}+a(\varepsilon_{P,f}-1)=0$
    \vspace{4pt}
    
   \STATE  Update parameters: 
   $\left\{
   \begin{array}{ll}
   \vec{r}=(r^{2C_{P}+1},r^{C_{P}+C_{N}}+r^{C_{P}+C_{N}}),\\
   s=\vec{c}\cdot\vec{r}.
   \end{array} \right.$\\
   
   \vspace{2pt}
   \STATE Keep $h_{f}(i)$ as round $t$ solution.
   \vspace{4pt}
   \ENDIF
   \vspace{4pt}
   \ENDIF}
}
   
   \vspace{4pt}
   \ENDFOR
   \vspace{4pt}
   \STATE $\star$ Calculate goodness parameter: $\alpha_{t}=\log{(r)}$
   \vspace{4pt}
   \STATE $\star$ Update weights subdistributions: 
   $\left\{
   \begin{array}{ll}
   D_{P}(i) \leftarrow D_{P}(i)\exp(-C_{P}\alpha_{t}h_{t}(i)),\\
   D_{N}(i) \leftarrow D_{N}(i)\exp(C_{N}\alpha_{t}h_{t}(i)).
   \end{array} \right.$\\
   \vspace{4pt}
   \ENDFOR
   \vspace{4pt}
   \STATE {\bfseries Final Classifier:}\\ 
   \vspace{4pt}
   $H(\mathbf{x})=\mathrm{sign}\left( \sum_{t=1}^{T}\alpha_{t}h_{t}(\mathbf{x})\right)$
\end{algorithmic}
\end{algorithm}

	\begin{algorithm}[p]
   \caption{Cost-Generalized AdaBoost}
   \label{adbg_algorithm}
   \scriptsize
   \begin{algorithmic}
   \STATE {\bfseries Input:} \\Training set of $n$ examples: $(\mathbf{x_i}, y_i)$, where $y_{i}=$
   $\left\{
   \begin{array}{ll}
   1 & \mbox{if $1 \leq i \leq m$} \text{  (Positives)},\\
   -1 & \mbox{if $m < i \leq n$} \text{  (Negatives)}.
   \end{array} \right.$\\    
   Pool of $F$ weak classifiers: $h_{f}(\mathbf{x})$\\ 
   Number of rounds: $T$\\
   $\star$ Two initial weight distributions over positives and negatives: $D_{+}(i)$ and $D_{-}(i)$\\ 
   $\star$ Cost parameters: $C_{P}$, $C_{N} \in \mathbb{R}^{+}$\\
   
   \vspace{4pt}
   \STATE {\bfseries Initialize:} \\ $\star$ Global weight distribution: $D(i)=$
   $\left\{
   \begin{array}{ll}
   \frac{C_{P}}{C_{P}+C_{N}}D_{+}(i) & \mbox{if $1 \leq i \leq m$},\\  			   
   \frac{C_{N}}{C_{P}+C_{N}}D_{-}(i) & \mbox{if $m < i \leq n$}.
   \end{array} \right.$\\
   
   \vspace{4pt}
   \STATE {\bfseries Iterate:} \\
   \FOR{$t=1$ {\bfseries to} $T$}
   \FOR{$f=1$ {\bfseries to} $F$}
   \vspace{4pt}
   \STATE Compute the weighted error of the $f^{th}$ weak classifier, $\epsilon_{t,f}= \sum_{i=1}^{n} D(i) \llbracket h_{f}(\mathbf{x}_{i}) \neq y_{i}\rrbracket$
   \vspace{4pt}
   \ENDFOR
   \vspace{4pt}
   \STATE Select the weak learner $h_t(\mathbf{x})$ of smallest error, $\epsilon_{t}=\underset{f}{\operatorname{arg min}} \left[\epsilon_{t,f}\right]$
   \STATE Compute the goodness parameter of the selected classifier, $\alpha_t=\frac{1}{2}\log\left(\frac{1-\epsilon_{t}}{\epsilon_{t}}\right)$
   \STATE Update weights: $D(i) \leftarrow \frac{D(i)\exp\left(-\alpha_{t}y_{i}h_{t}\left(\mathbf{x}_i\right)\right)}{\sum_{i=1}^{n}D(i)\exp\left(-\alpha_{t}y_{i}h_{t}\left(\mathbf{x}_i\right)\right)}$
	 \vspace{4pt}
   \ENDFOR
   \vspace{4pt}
   \STATE {\bfseries Final Classifier:}\\
   \vspace{4pt}
    $H(\mathbf{x})=\mathrm{sign}\left(\sum_{t=1}^{T}\alpha_{t}h_{t}(\mathbf{x})\right)$
\end{algorithmic}
\end{algorithm}

\clearpage

\section{Other Cost Scenarios}
\label{app:cost_scen}

\begin{figure}[p]
\centering
\includegraphics[width=0.9\columnwidth]{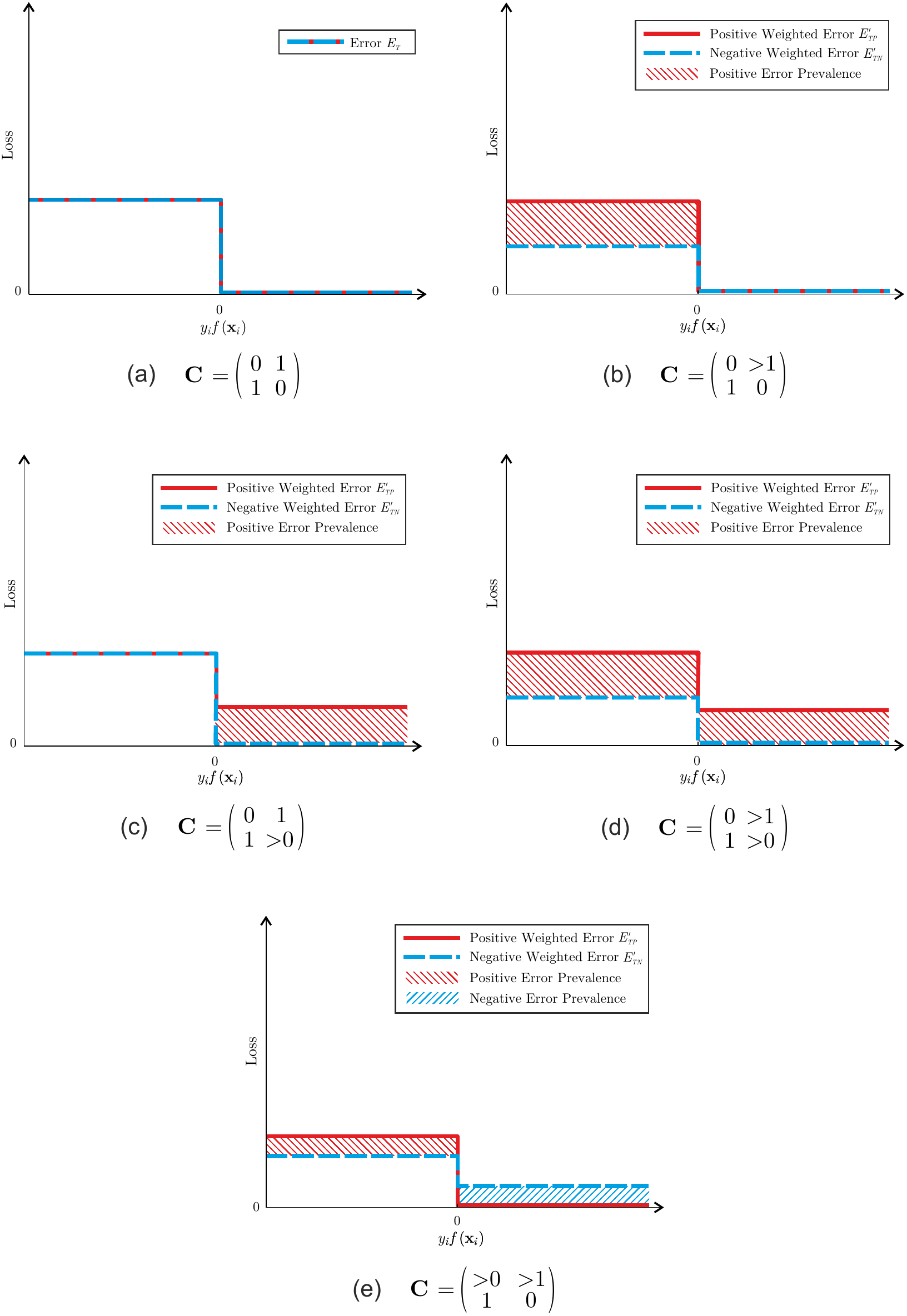}
\caption{Different cost scenarios [Symmetry (a), Error Asymmetry (b), Correct Classification Asymmetry (c) Dual Asymmetry (d) and Reversed Dual Asymmetry (e)] with their corresponding cost matrices.}
\label{cost_scen_figs} 
\end{figure}

As we have analyzed so far, the most common detection problem is that in which cost coefficients are null for correct decisions ($c_{nn}=c_{pp}=0$) but non-zero for mistakes  ($c_{np},c_{pn}>0$). Thus, we have distinguished between two ``usual'' scenarios:

\begin{itemize}
\item \emph{Cost-Insensitive (Symmetry)}: Regardless of the class, all mistakes have the same cost ${c}_{np}={c}_{pn}$ (Figure \ref{cost_scen_figs}a).
\item \emph{Cost-Sensitive (Error Asymmetry)}: Mistakes in positives are costlier than mistakes in negatives ${c}_{np}>{c}_{pn}$ (Figure \ref{cost_scen_figs}b).
\end{itemize}

However, ``reasonableness conditions'' ($c_{nn}<c_{np}$ and $c_{pp}<c_{pn}$) \citep{Elkan01} commented in Section \ref{sec:CSvar}, still allow other possible scenarios depending on how costs are defined:

\begin{itemize}
\item \emph{Correct Classification Asymmetry}: All mistakes have the same cost ${c}_{np}={c}_{pn}$, but correct decisions on positives are costlier than on negatives $c_{pp}>c_{nn}$ (Figure \ref{cost_scen_figs}c).
\item \emph{Dual Asymmetry}: Correct and wrong decisions on positives are costlier than correct and wrong decisions on negatives respectively, ${c}_{pp}>{c}_{nn}, {c}_{np}>{c}_{pn}$ (Figure \ref{cost_scen_figs}d).
\item \emph{Reversed Dual Asymmetry}: Mistakes on positives are costlier than on negatives ${c}_{np}>{c}_{pn}$, while correct decisions are costlier on negatives than on positives ${c}_{nn}>{c}_{pp}$ (see Figure \ref{cost_scen_figs}e).
\end{itemize}

In all these cases we have supposed, without loss of generality, that the cost of mistakes in positives is always greater or equal than the cost of mistakes in negatives ($c_{np}\geq c_{pn}>0$), since the opposite case can be modeled just by swapping labels.

\begin{figure}[p]
\centering
\includegraphics[width=0.9\columnwidth]{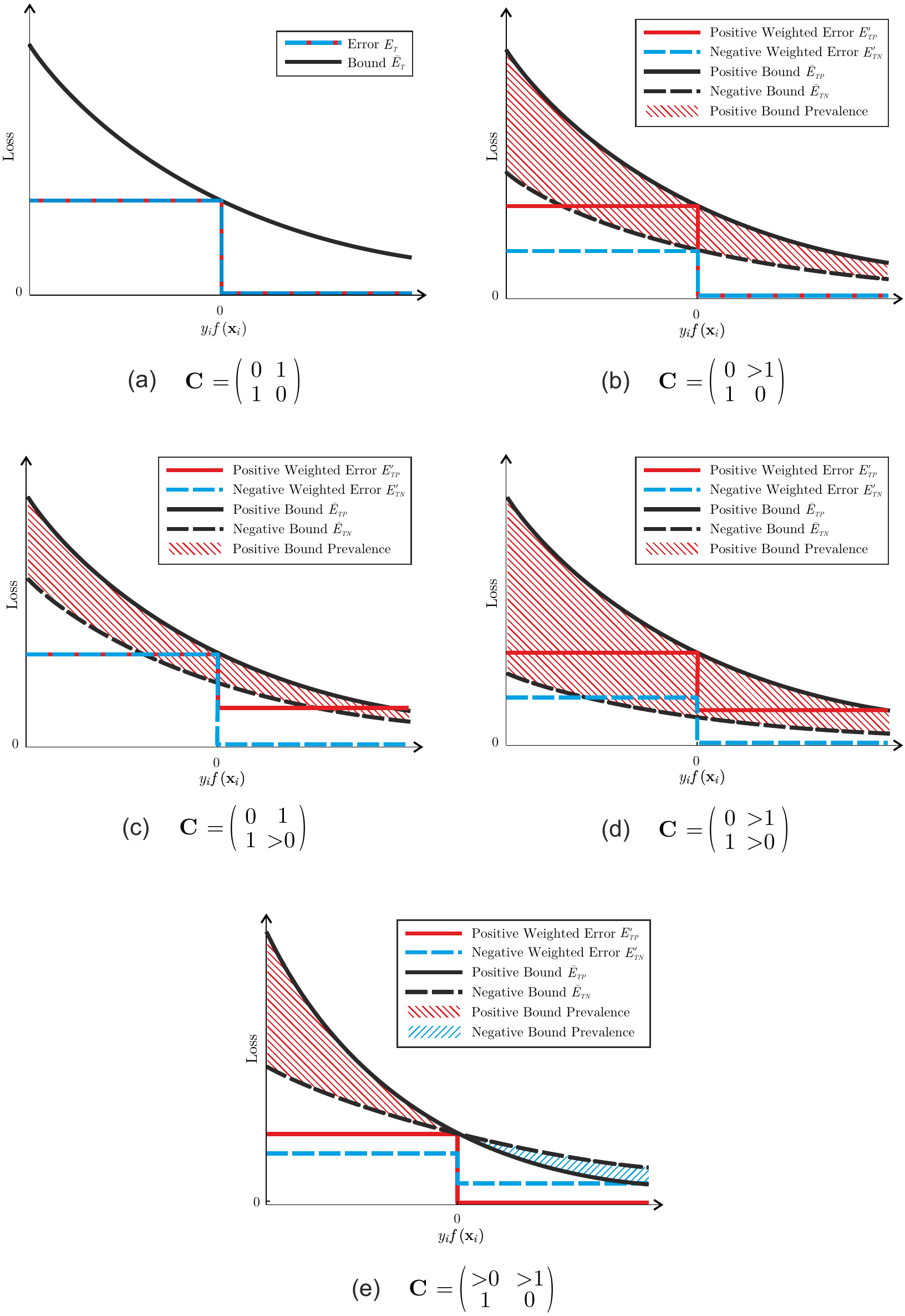}
\caption{Boosting models applied to the cost scenarios in Figure \ref{cost_scen_figs} [AdaBoost for Symmetry (a), Cost-Generalized AdaBoost for Error Asymmetry (b), Cost-Generalized AdaBoost for Correct Classification Asymmetry (c), Cost-Generalized AdaBoost for Dual Asymmetry (d) and Cost-Sensitive AdaBoost for Reversed Dual Asymmetry (d)].}
\label{cost_scen_boosting_figs} 
\end{figure}

As we have seen, classical AdaBoost, with its standard exponential bound, is aimed to deal with the cost-insensitive case (Figure \ref{cost_scen_boosting_figs}a), while Cost-Generalized AdaBoost seems to be best suited than Cost-Sensitive AdaBoost for the standard cost-sensitive (error asymmetric) scenario (Figure \ref{cost_scen_boosting_figs}b). But, what would happen in the other asymmetric scenarios? Which of the two theoretical variants is best suited for each case?

\begin{itemize}
\item \emph{Correct Classification Asymmetry}: The positive class is prevalent for positive performance scores while no class is prevalent for negative ones. Cost-Generalized AdaBoost, with a smoothed asymmetry, seems to be the most suitable scheme (Figure \ref{cost_scen_boosting_figs}c).
\item \emph{Dual Asymmetry}: The positive class is hegemonic throughout the whole performance score space, and the only difference of being on either side of the success boundary is the cost ratio between the two classes. Cost-Generalized AdaBoost, in this case with a more pronounced asymmetry, may be the most appropriate scheme (Figure \ref{cost_scen_boosting_figs}d).
\item \emph{Reversed Dual Asymmetry}: The costlier class changes depending on being mistaken or not. In this case, Cost-Sensitive AdaBoost, taking advantage of its class-prevalence reversal, seems to be the most suitable model for the problem (Figure \ref{cost_scen_boosting_figs}e).
\end{itemize}

Up to this point we have assumed that cost coefficients are constant, so all the examples belonging to the same class have the same associated cost. However, a cost matrix with variable coefficients would entail that different examples of the same class may have different costs. In general terms, cost requirements of any classification problem can be split into two levels: a \emph{Class-Level} regarding the cost ratio between classes (the global emphasis given to each class), and an \emph{Example-Level} regarding the cost distribution within a given class. When cost coefficients are constant, class-level is the only kind of asymmetry involved in the problem, but when costs are variable both levels can be present.

As analyzed in \citep{LandesaAlba12}, this asymmetry breakdown can be immediately mapped to the Cost-Generalized AdaBoost framework by only defining a set of parameters from the initial weight distribution $D(i)$ given to the algorithm:

\vspace{3pt}
\begin{itemize}
\item \emph{Asymmetry}:
\begin{equation}
\label{asym_eqn}
\sum_{i=1}^{m} D(i) = \gamma\in (0,1)
\end{equation}
\item \emph{Class-conditional weight distributions}:
\begin{gather}
\label{pos_weight_ini_eqn}
D_{P}(i)=
\frac{D(i)}{\gamma}, \quad \textrm{for } i=1,\ldots,m\\
\label{neg_weight_ini_eqn}
D_{N}(i)=
\frac{D(i)}{1-\gamma}, \quad \textrm{for } i=m+1,\ldots,n
\end{gather}
\end{itemize}
\vspace{3pt}

\begin{figure}[!htb]
\centering
\includegraphics[width=0.9\columnwidth]{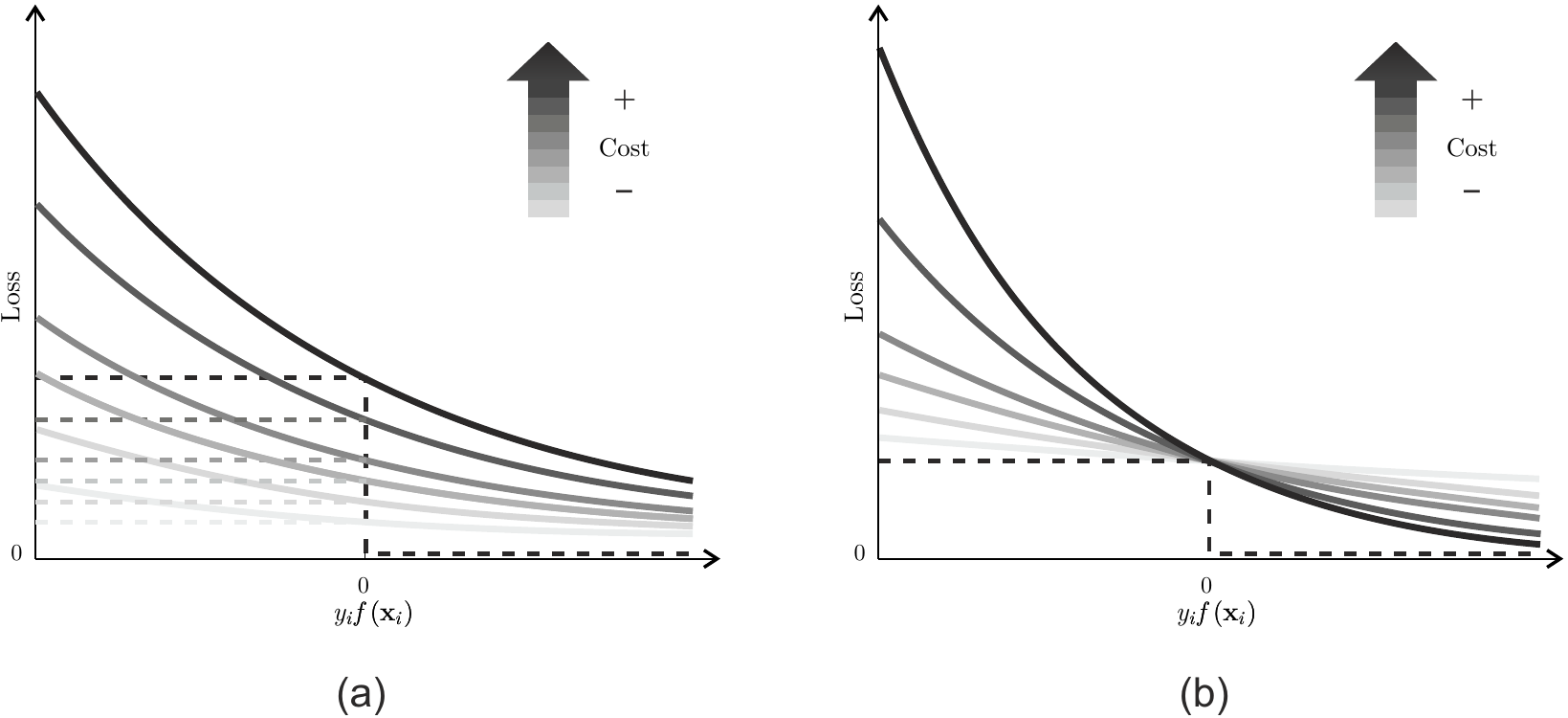}
\caption{Variable cost coefficients mapped to different initial weights on Cost-Generalized AdaBoost (a), and to different exponential bases in Cost-Sensitive AdaBoost(b).}
\label{var_bound_fig} 
\end{figure}

While $\gamma$ quantifies the class-level global asymmetry of the problem, class-conditional weight distributions $D_P(i)$ and $D_N(i)$ describe how some examples are emphasized within each class (example-level). Hence, initial weights $D(i)$ are coupling both cost levels, and determine the specific exponential bound applied to each example throughout the minimization procedure (\refeq{cgb_bound2}). As a result, by simply defining a proper weight initialization scheme, Cost-Generalized AdaBoost is able to model any class-level or element-level asymmetric cost scenario (without class prevalence reversal) and it also preserves the same computational complexity for all cases. Figure \ref{var_bound_fig}a illustrates the effect of mapping variable cost coefficients to different weights in Cost-Generalized AdaBoost.

\begin{equation}
\label{cgb_bound2}
\begin{split}
E_{T} &= \sum_{i=1}^{n} D(i) \llbracket H(\mathbf{x}_{i}) \neq y_{i}\rrbracket \leq \sum_{i=1}^{n} D(i) \exp \left( -y_{i} f(\mathbf{x}_{i}) \right) = \tilde{E}_{T}
\end{split}
\end{equation}

In case of having variable cost coefficients and class prevalence reversal (Reversed Dual Asymmetry), a modification of Cost-Sensitive AdaBoost or AdaBoostDB is needed. Such a variation would require distinct exponent modulators for each cost, so the resulting global error bound would be modeled (\refeq{csb_bound2}) as a sequence of exponential factors with different bases related to each different cost. However, minimization of this bound will be increasingly complex depending on the number of distinct ``discrete costs'' we have  in the training set, to the point that the process may end up being unfeasible: as we will see in the accompanying paper of the series \citep{LandesaAlba??b}, just passing from one base (Standard AdaBoost and Cost-Generalized AdaBoost) to two different bases (Cost-Sensitive AdaBoost/AdaBoostDB with constant coefficients) implies that training time becomes 18 times longer, on average, even using the most efficient of the two alternatives\footnote{As shown in \citep{LandesaAlba13}, AdaBoostDB is, on average, 200 times faster than Cost-Sensitive AdaBoost}. The graphical behavior of Cost-Sensitive AdaBoost when variable cost coefficients are mapped to different exponential bases can be visualized in Figure \ref{var_bound_fig}b.

\begin{equation}
\label{csb_bound2}
\begin{split}
E_{T} &= \sum_{i=1}^{m_1} \frac{1}{n} \llbracket H(\mathbf{x}_{i}) \neq y_{i}\rrbracket + \sum_{i=m_1+1}^{m_2} \frac{1}{n} \llbracket H(\mathbf{x}_{i}) \neq y_{i}\rrbracket + \ldots + \sum_{i=m_k+1}^{n} \frac{1}{n} \llbracket H(\mathbf{x}_{i}) \neq y_{i}\rrbracket \\
& \leq \sum_{i=1}^{m_1} \frac{1}{n} \beta_1^{-y_{i} f(\mathbf{x}_{i})} + \sum_{i=m_1+1}^{m_2} \frac{1}{n} \beta_2^{-y_{i} f(\mathbf{x}_{i})} + \ldots + \sum_{i=m_k+1}^{n} \frac{1}{n} \beta_k^{-y_{i} f(\mathbf{x}_{i})} = \tilde{E}_{T}
\end{split}
\end{equation}

Table \ref{asymm_mapping_tab} summarizes all these conclusions. As can be seen, Cost-Generalized AdaBoost dominates most of the possible cost scenarios, including the most common ones (symmetry and standard asymmetry\footnote{As analyzed in \citep{LandesaAlba12} AdaBoost and Cost-Generalized AdaBoost can be actually considered as one algorithm, preserving both theoretical properties and computational complexity for any cost requirements.}), it is able to model both constant and variable cost coefficients and it also preserves the same computational complexity for all cases.

\begin{table}[!htb]
\scriptsize
\centering
\begin{adjustbox}{max width=\textwidth}
{
\renewcommand{\arraystretch}{0.9}
\newcolumntype{S}{>{\centering\arraybackslash} m{.25\columnwidth} }

\vspace{8pt}
\begin{tabular}[c]{|S|S|S|S|}
\hline
\multirow{2}{*}{\textbf{ Scenario }} & \multirow{2}{*}{\textbf{ Cost Matrix }} &\multicolumn{2}{|c|}{\textbf{Level of Asymmetry}}\\ \cline{3-4}
& & \textbf{Class-Level} & \textbf{Example-Level}\\ \hline

\textbf{Symmetry} & 
\begin{tabular}{c}
\\
$\mathbf{C}=\left(
\begin{array} {c c}
0 & 1\\
1 & 0\\
\end{array}
\right)$\\ \\
\end{tabular} 
& AdaBoost (Cost-Generalized AdaBoost)
& Cost-Generalized AdaBoost\\ \hline
  
\textbf{Standard Asymmetry} & 
\begin{tabular}{c}
\\
$\mathbf{C}=\left(
\begin{array} {c c}
0 & >1\\
1 & 0\\
\end{array}
\right)$\\ \\
\end{tabular} 
& Cost-Generalized AdaBoost
& Cost-Generalized AdaBoost\\ \hline

\textbf{Unbalanced Asymmetry} & 
\begin{tabular}{c}
\\
$\mathbf{C}=\left(
\begin{array} {c c}
0 & >1\\
1 & >0\\
\end{array}
\right)$\\ \\
\end{tabular} 
& Cost-Generalized AdaBoost 
& Cost-Generalized AdaBoost\\ \hline

\textbf{Reversed Asymmetry} & 
\begin{tabular}{c}
\\
$\mathbf{C}=\left(
\begin{array} {c c}
>0 & >1\\
1 & 0\\
\end{array}
\right)$\\ \\
\end{tabular}
& Cost-Sensitive AdaBoost 
& Cost-Sensitive AdaBoost (increasingly complex)\\ \hline
\end{tabular}
}
\end{adjustbox}
\caption{Summary of the proposed mapping between AdaBoost algorithms and the different asymmetric scenarios. }
\label{asymm_mapping_tab}
\end{table}

\clearpage

\section{Comments on Figures \ref{counter_example_1_fig} and \ref{counter_example_2_fig}} 
\label{app:comments_figures}

If we analize Figures \ref{counter_example_1_fig}c and \ref{counter_example_2_fig}c we will see that, before error profiles start to iteratively evolve, both classifiers have an initial period of ``flat'' performance progress.

\begin{itemize}

\item  In the case of Figure \ref{counter_example_1_fig}c the classifier obtained after the first round yields null classification error in positives and 0.7 error in negatives, maintaining the same performance until round 6.
\item In Figure \ref{counter_example_2_fig}c, the first round classifier gets null error in positives and total error in negatives (it is  an ``all-positives'' classifier), keeping this global performance unchanged until round 9.

\end{itemize}

What is happening during these seemingly ``stubborn'' rounds? Does an ``all-positives'' or an ``all-negatives'' weak classifier makes sense inside the AdaBoost framework?

In the first round the weak learner selects the best weak classifier for the initial weight distribution. Bearing in mind that initial weights define the overall desired asymmetry for the whole problem \citep{LandesaAlba12}, the first learning round is actually searching a weak classifier ``as if'' it was going to be the only one in the ensemble. Depending on the topology of the classes, the pool of weak classifiers and the desired asymmetry, it could happen that the best single weak classifier dealing with the problem is an ``all-positives'' or an ``all-negatives'' one. As we can easily see in Figure \ref{counter_example_2_fig}, due to the spatial distribution of both classes, their relative costs and the weak classifiers we have used (stumps in the linear bidimensional space), the best possible single weak classifier is just an ``all-positives'' one. This effect can be interpreted as a \emph{global asymmetry adjustment}, fitting the a-priori probability of each class and smoothing the weight distribution, so the subsequent training rounds can select new weak classifiers to jointly build a more accurate strong classification. Note that classifiers depicted in Figures \ref{counter_example_1_fig}a and \ref{counter_example_2_fig}a are the first four weak classifiers obtained by AdaBoost excluding the ``all-positives'' or ``all-negatives'' weak-classifier rounds.

Also depending on the topology of the problem and the desired asymmetry, once the first round is trained, it is possible that the best weak classifier of the next iteration has not ``enough goodness'' ($\alpha_t$) to change the decision boundaries of the strong classifier. In that case, even though the predictor will evolve according to the incorporated weak hypothesis, no changes in the performance of the strong classifier will be perceived. This situation may be repeated for several iterations, and it is the responsible of the aforementioned initial \emph{flat sections} in Figures \ref{counter_example_1_fig}c and \ref{counter_example_2_fig}c. The key is that no single weak classifier has enough complexity to effectively contribute to change itself the strong classifier, so several consecutive weak hypothesis must be gathered to jointly achieve enough performance to change the decision boundaries.

These two phenomena (``global asymmetry adjustment rounds'' and ``flat sections') are not exclusive of the first round, and can be found, with the same meaning, at any other different point of the training process different (e.g. rounds 13 to 18 in Figure \ref{counter_example_2_fig}c). 

\clearpage



\section{Cost-Sensitive Boosting Extensions and Weight Initialization} 
\label{app:weight}

Original (Discrete) AdaBoost can be generalized to deal with weak hypotheses that, instead of being binary $\{-1,1\}$, are defined over a continuous range in $\mathbb{R}$ \citep{SchapireSinger99}. This extension of the algorithm is usually known as Real AdaBoost \citep{Friedman00}, and it is based on the minimization of the same exponential loss as in the discrete case. As a consequence, for cost-sensitive purposes, the same loss modification and weight initialization strategies used in Cost-Generalized AdaBoost \citep{LandesaAlba12} can be applied, preserving again all the theoretical guarantees of the symmetric Real AdaBoost version. 

Besides Cost-Sensitive AdaBoost, in \citep{MasnadiVasconcelos11} a Cost-Sensitive RealBoost version is also proposed. For its derivation, the authors use the same exponential loss (\refeq{app:csloss_eqn1}) as in the discrete case, with asymmetry embedded in the exponents, so the cost-proportionate weight initialization procedure (linked to a direct modulation of the exponential components) is again discarded.

\begin{equation}
\label{app:csloss_eqn1}
J(f(\mathbf{x}))=\E\left[\llbracket y=1 \rrbracket \mathrm{e}^{-C_{P}f(\mathbf{x})} + \llbracket y=-1 \rrbracket \mathrm{e}^{C_{N}f(\mathbf{x})}\right]
\end{equation}

The Cost-Sensitive Boosting framework, also includes a third algorithm: Cost-Sensitive LogitBoost. Unlike the previous cases, the original (cost-insensitive) LogitBoost algorithm \citep{Friedman00} is not based on minimizing an exponential loss, but on maximizing a Bernoulli log-likelihood (\refeq{app:logloss_exp}):
\begin{equation}
\label{app:logloss_exp}
l(f(\mathbf{x}))=\E\left[y'\log(\mathrm{p}(\mathbf{x}))+(1-y')\log(1-\mathrm{p}(\mathbf{x}))\right]
\end{equation}
where
\begin{gather}
y'=\frac{y+1}{2}\\
\Prob(y'=1|\mathbf{x})=\mathrm{p}(\mathbf{x})=\frac{\mathrm{e}^{F(\mathbf{x})}}{\mathrm{e}^{F(\mathbf{x})}+\mathrm{e}^{-F(\mathbf{x})}}
\end{gather}

Then, to derive their Cost-Sensitive LogitBoost proposal, Masnadi-Shirazi and Vasconcelos define the probability of $y'=1$ as follows
\begin{equation}
\mathrm{p}(\mathbf{x})=\frac{\mathrm{e}^{\lambda f(\mathbf{x})+\eta}}{\mathrm{e}^{\lambda f(\mathbf{x})+\eta}+\mathrm{e}^{-\lambda f(\mathbf{x})-\eta}}
\end{equation}
\begin{gather}
\gamma=\frac{C_{1}+C_{2}}{2}\\
 \eta=\frac{1}{2}\log\left(\frac{C_{2}}{C_{1}}\right)
\end{gather}

If we normalize cost factors by $2/(C_{1}+C_{2})$ in the expressions above (note that the only relevant issue is the cost proportion, not their actual value), it is easy to see that, such a definition of $\mathrm{p}(\mathbf{x})$ is, in fact, equivalent to (\refeq{prob_logit})

\begin{equation}
\label{prob_logit}
\mathrm{p}(\mathbf{x})=\frac{C_{2}\mathrm{e}^{f(\mathbf{x})}}{C_{2}\mathrm{e}^{f(\mathbf{x})}+C_{1}\mathrm{e}^{-f(\mathbf{x})}}
\end{equation}

As a result, the optimization strategy inside Cost-Sensitive LogitBoost, is actually based on embedding asymmetry by modulation of the exponential terms. Curiously, such an approach is just the same mechanism that, in the same work \citep{MasnadiVasconcelos11}, was rejected for AdaBoost and  RealBoost. Moreover, this strategy eventually becomes a different \emph{initial weight distribution} for the first weighted least-squares regression iteration of LogitBoost, in clear analogy with the mechanism in which Cost-Generalized AdaBoost \citep{LandesaAlba12} is based.

\end{appendices}